\pdfoutput=1

\documentclass[11pt]{article}

\usepackage[final]{acl}

\usepackage{times}
\usepackage{latexsym}
\usepackage{enumitem}
\usepackage{amssymb}
\usepackage{booktabs}
\usepackage{float}
\usepackage{placeins}
\usepackage{amsmath}
\usepackage{multirow}
\usepackage{parskip}
\usepackage{color,soul}

\usepackage[T1]{fontenc}

\usepackage[utf8]{inputenc}

\usepackage{microtype}

\usepackage{inconsolata}

\usepackage{graphicx}

\title{Affect Recognition in Conversations Using Large Language Models}

\author{Shutong Feng$^{1}$, Guangzhi Sun$^{2}$, Nurul Lubis$^{1}$, Wen Wu$^{2}$, Chao Zhang$^{3}$, Milica Ga\v{s}i\'{c}$^{1}$\\
  $^{1}$Heinrich Heine University Düsseldorf, Germany\\
  $^{2}$University of Cambridge, UK\\
  $^{3}$Tsinghua University, China\\
  \texttt{\{fengs,lubis,gasic\}@hhu.de$^{1}$,\{gs534,ww368\}@cam.ac.uk$^{2}$,cz277@tsinghua.edu.cn$^{3}$}}

\begin{document}
\maketitle
\begin{abstract}
Affect recognition, encompassing emotions, moods, and feelings, plays a pivotal role in human communication. In the realm of conversational artificial intelligence, the ability to discern and respond to human affective cues is a critical factor for creating engaging and empathetic interactions. This study investigates the capacity of large language models (LLMs) to recognise human affect in conversations, with a focus on both open-domain chit-chat dialogues and task-oriented dialogues. Leveraging three diverse datasets, namely IEMOCAP \citep{iemocap}, EmoWOZ \citep{feng-etal-2022-emowoz}, and DAIC-WOZ \citep{gratch-etal-2014-distress}, covering a spectrum of dialogues from casual conversations to clinical interviews, we evaluate and compare LLMs' performance in affect recognition. Our investigation explores the zero-shot and few-shot capabilities of LLMs through in-context learning as well as their model capacities through task-specific fine-tuning. Additionally, this study takes into account the potential impact of automatic speech recognition errors on LLM predictions. With this work, we aim to shed light on the extent to which LLMs can replicate human-like affect recognition capabilities in conversations. 

\end{abstract}

\section{Introduction}

Affect refers to the broad range of subjective experiences related to emotions, moods, and feelings \citep{1981-25062-001}. It encompasses the various ways individuals perceive, experience, and express their emotional states and is an essential aspect of human experience and communication \citep{gross_2002}.

The ability to recognise human affect is an important ability of conversational artificial intelligence (AI, \citealt{MAYER1999267}). It empowers the dialogue agent to go beyond mere information exchange and engage users on an emotional level. By leveraging affect recognition techniques, they can discern the emotional nuances in user inputs, including sentiment, mood, and subtle cues like sarcasm or frustration \citep{Picard97}. This capability allows the system to respond with greater sensitivity, empathy, and relevance, leading to more meaningful and satisfying interactions \citep{4468714}. 

Large language models (LLMs) have demonstrated promising performance in many tasks \citep{open-llm-leaderboard}. They have also shown promising capability in adapting to new tasks via prompting \citep{chatgpt-dst,sun2023speechbasedslotfillingusing}, in-context learning (ICL, \citealt{chatgpt-emotional-dialogue}), as well as task-specific fine-tuning \citep{alpaca}. With the advancement in LLMs, it is possible to use LLMs as the backend of dialogue systems \citep{chatgpt,gpt4,llama2}. This brings up the question: can LLMs recognise human affects in conversations in a similar capacity as human beings?

In the context of conversational AI, dialogues can be broadly categorised into two classes: 1) chit-chat or open-domain dialogues where users interact with the system for entertainment and engagement, and 2) task-oriented dialogues (ToDs) where users converse with the system for specific goals \citep{jurafsky}. Under ToDs, depending on the type of user goals, dialogues can be further grouped as information-retrieval, medical consultations, education, and many more. 

Regarding the affective information in conversations, we are particularly interested in the following: (1) categorical emotion classes from generic emotion models such as ``basic emotions'' proposed by \citet{ekman-basic}, (2) custom categorical emotion classes defined for a particular context, such as the emotion labels defined by~\citet{feng-etal-2022-emowoz} to encode task performance simultaneously in ToDs, and (3) depression, a medical illness that negatively affects how a person feels, thinks and acts, and causes feelings of sadness and/or a loss of interest in activities the person once enjoyed~\citep{depression-definition}. 

The emergence of LLMs has signified a shift of paradigm from training small models for one specific task to large models for multiple tasks. Therefore, in this work, we investigate the affect recognition ability of a range of LLMs on vastly different types of dialogues and labels\footnote{The code can be found at \url{https://gitlab.cs.uni-duesseldorf.de/general/dsml/llm4erc-public/}} to ascertain the validity of this direction. Specifically,

\begin{itemize}[leftmargin=*,noitemsep]
 \item We evaluated and compared the ability of a range of LLMs to recognise human affect under different dialogue set-ups (chit-chat dialogues and ToDs) and recognition targets (emotion classes and binary depression diagnosis). We used the following datasets: IEMOCAP \citep{iemocap}, EmoWOZ \citep{feng-etal-2022-emowoz}, and DAIC-WOZ \citep{gratch-etal-2014-distress}.
 
 \item We investigated into LLMs' zero-shot and few-shot capabilities through an array of ICL set-ups as well as their model capacities through task-specific fine-tuning. 
 
 \item We considered text-based LLMs as a part of spoken dialogues systems. Therefore, we also experimented with inputs containing automatic speech recognition (ASR) errors to investigate the potential influence of ASR errors on LLM predictions.
\end{itemize}

\section{Related Work}
\subsection{LLM}
Large Language Model~(LLM) refers to a type of pre-trained models designed for natural language processing tasks. LLMs are characterised by their enormous number of model parameters and extensive training data.

Some well-known examples of LLMs include OpenAI GPT family models \citep{gpt2,chatgpt,gpt4}, which can have billions or even trillions of model parameters. Examples of open-source text-based foundation models include the LLaMA family \citep{llama,llama2,llama3modelcard} and their corresponding chat-optimised models.
These models have demonstrated remarkable abilities in various natural language understanding and generation tasks, including text completion, language translation, text summarisation, and even chatbot applications \citep{open-llm-leaderboard}. They also demonstrate ``emergent abilities'' such as few-shot prompting and chain-of-thought reasoning, which were not present in their smaller predecessors \citep{emergent}. While there are also multi-modal LLMs such as SALMONN \citep{tang2024salmonn}, these are at an earlier stage compared to uni-modal text-based LLMs, and it is still a common practice to use text-based LLMs as the text-processing backend, pipelined with other modules such as ASR and image generator for more complex applications.

\subsection{Affective Capabilities of LLMs}
With the growing attention on LLMs from the research community, there have been several works investigating the affective abilities of LLMs. \citet{llm-empathy} evaluated the empathy ability of LLMs by utilising the emotion appraisal theory from psychology. \citet{llm-eq} assessed the emotional intelligence of LLMs in terms of Emotional Quotient (EQ) scores. \citet{zhang2023sentiment} investigated how LLMs could be leveraged for a range of sentiment analysis tasks under zero-shot or few-shot learning set-ups. \citet{chatgpt-emotional-dialogue} investigated the emotional dialogue ability of ChatGPT through a range of understanding and generation tasks.
In our work, we focus on the affect recognition ability of text-based LLMs. Our investigation spans across different types of dialogues and model learning set-ups. We also consider real-world applications of LLMs and consider ASR-inferred noisy input to models.

\section{Methodology}

The ability of human-beings to recognise affect can be reflected in many ways.
Yet, being able to narrate what emotion has been expressed in the utterances of the other interlocutor is a straightforward and strong sign of such an ability. Therefore, we took LLMs' ability to verbalise the emotion given the dialogue context as a proxy to both qualitatively and quantitatively analyse LLMs' ability for affect recognition.

\subsection{Affect Recognition using LLMs}
\label{sec:}

\begin{figure*}[h]
    \centering
    \includegraphics[width=0.8\textwidth]{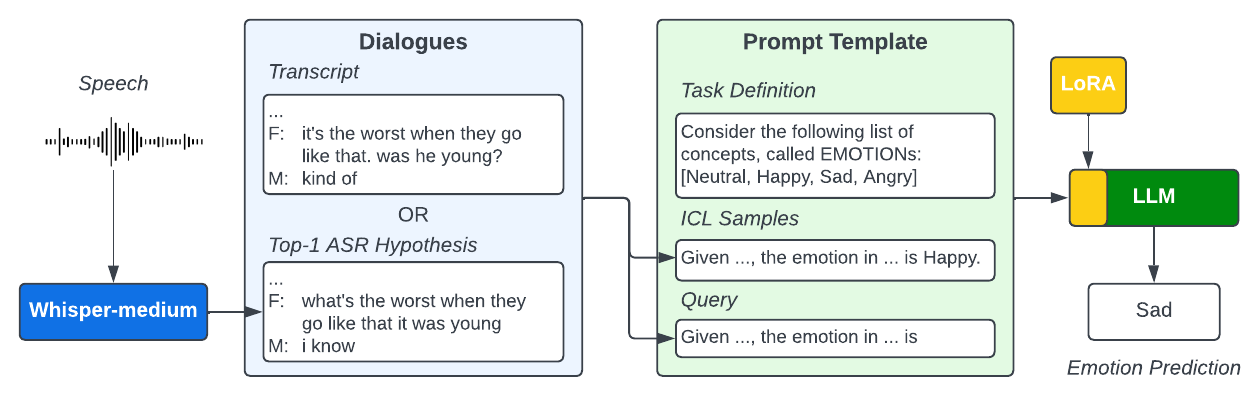}
    \caption{A flowchart illustrating the affect recognition pipeline using Whisper and LLM. The designed prompt comprises parts introduced in Table \ref{tab:prompt-design}. Low-rank adaptation (LoRA) is used for fine-tuning open-source LLMs.}
    \label{fig:llm-pipeline}
\end{figure*}

The pipeline for affect recognition using LLMs with the option to take speech as input is illustrated in Figure \ref{fig:llm-pipeline}. When using the speech input, a Whisper-medium model was used to transcribe the speech (see Section \ref{sec:asr-specifications} for details). The prompt is then constructed as designed and fed into the LLM to generate a text sequence. For open-source LLMs, we examined the probability of each class token and considered the one with the maximum probability as the final model prediction, as shown in Equation \ref{eqn:search}.
\begin{equation}
    \mathbf{W_L}* = \underset{\mathbf{W_L}}{\arg\max} P(\mathbf{W_L}|\mathbf{W}^P),
    \label{eqn:search}
\end{equation}
where $\mathbf{W_L}$ belongs to the set of pre-defined labels and $\mathbf{W}^P$ is the prompt token sequence. 

For commercial models, there is no access to logits of model outputs and model outputs do not always follow the format specified in the prompt. Therefore, we used regular expressions to derive the final prediction.

\subsection{Task-specific Fine-tuning}
\label{sec:task-specific-fine-tuning}
For efficient training of LLMs, we utilise low-rank adaptation (LoRA, \citealt{lora}) to accelerate the fine-tuning of LLMs while conserving memory. This is also a common approach for fine-tuning LLMs as seen in many existing works \citep{sun2023speechbasedslotfillingusing,zhao2024loraland310finetuned}.

LoRA hypothesises that the change in weights during model training has a low ``intrinsic rank''. Therefore, instead of directly updating the full-rank weight matrices of dense layers during training, LoRA optimises the low-rank decomposition matrices of those dense layers' changes while keeping the pre-trained weights frozen. Specifically, for a pre-trained weight matrix $\mathbf{W}_0 \in \mathbb{R}^{m \times n}$ from a particular attention block in a transformer-based LLM, its update $\Delta\mathbf{W}$ is constrained using a low-rank decomposition of the update as following:
\begin{equation}
\label{eqn:lora}
    \mathbf{W}_0 + \Delta\mathbf{W} = \mathbf{W} + \mathbf{AB}
\end{equation}
where matrices $\mathbf{A}\in \mathbb{R}^{m \times r}$ and $\mathbf{B}\in \mathbb{R}^{r \times n}$ contain trainable parameters and $r \leq \textrm{min}(m,n)$. The pre-trained parameters in $\mathbf{W}_0$ are fixed. When $r$ is set to a much smaller value than the dimensions of $\mathbf{W}_0$, the number of trainable parameters will be greatly reduced. This leads to greater training efficiency, less memory requirement, and a lower chance of over-fitting. Following \citet{lora}, we apply LoRA to the projection matrices of the self-attention layers of transformer-based LLMs.

LLMs are trained to predict the next token in the sequence (the label tokens), given the previous tokens (the designed prompt). During training, the input tokens are fed into the model, and the model predicted the probability distribution of the next token. The cross-entropy loss is calculated from the model prediction and the target token. 

With LoRA, it takes roughly 30GB memory and 4 hours to train one epoch on the entire EmoWOZ training set using an Nvidia A100 40GB graphics card.

\section{Experimental Setup}
\subsection{Datasets and Evaluation}
\label{sec:datasets}
The \textbf{IEMOCAP} dataset \citep{iemocap} is a multi-modal corpus designed for Emotion Recognition in Conversations (ERC) task in chit-chat or open-domain dialogues. It comprises 151 dialogues, containing 10,039 utterances from 10 distinct speakers involved in 5 dyadic conversational sessions. Each utterance underwent annotation by a minimum of three annotators, who assigned one of nine emotion classes, including \textit{sad}, \textit{neutral}, \textit{angry}, \textit{happy}, \textit{excited}, \textit{frustrated}, \textit{surprised}, \textit{fearful}, \textit{disgusted}. Annotators could also assign multiple emotions or use the category ``other'' if the perceived emotion did not match the predefined options. Final labels were determined via majority voting.

Given the absence of an official train-test split, we adopt leave-one-session-out 5-fold cross-validation approach and average the results. 
Our methodology aligns with the common practices, as discussed by \citet{iemocap-sota}, to consider two label sets: \textbf{4-way}: \textit{Sad}, \textit{Neutral}, \textit{Angry}, and \textit{Happy}; \textbf{5-way}: \textit{Sad}, \textit{Neutral}, \textit{Angry}, \textit{Happy}, and everything else as \textit{Other}. In both set-ups, \textit{Excited} is merged with \textit{Happy}. 

Emotion recognition is performed for every speaker utterance. We report the weighted accuracy (WA) and unweighted accuracy (UA) for both label sets.

\textbf{EmoWOZ} \citep{feng-etal-2022-emowoz} is a text-based ERC corpus built for emotion recognition in ToDs. It comprises 10,438 human-human dialogues from the entire MultiWOZ dataset \citep{budzianowski-etal-2018-multiwoz}, as well as 1,000 human-machine dialogues in the same set of domains. It encompasses seven distinct user emotions, namely: \textit{Neutral, Fearful, Dissatisfied, Apologetic, Abusive, Excited,} and \textit{Satisfied}. These emotion labels are designed to encode the task performance. Specifically, each emotion is defined as a valence reaction to certain elicitor under certain conduct. For example, \textit{Dissatisfied} is defined as a negative emotion elicited by the system expressed in a neutral or polite conduct. 

Emotion recognition is performed for each \textbf{user} utterance. For existing benchmarks reported in \citet{feng-etal-2022-emowoz}, neutral class was excluded from calculating the metrics because they take up more than 70\% of the labels in EmoWOZ. To have a direct comparison, we report macro-averaged F1 and weighted average F1 excluding neutral. We include the F1, precision, and recall of the neutral class in Table \ref{tab:emowoz-full} of Appendix \ref{sec:detailed-results}.

\textbf{DAIC-WOZ} \citep{gratch-etal-2014-distress} is a speech-based corpus for depression detection and analysis. It includes the Patient Health Questionnaire-8 (PHQ-8, \citealp{Kroenke2008-qm}) scores of 193 clinical interviews, with 35 (12 are labelled depressed) interviews in the development set and 47 (14 are labelled depressed) in the test set. The PHQ-8 score ranges from 0 to 24 and quantifies the severity of the patient's depressive symptoms. 

For evaluation metrics, we follow the criteria established by the Audio/Visual Emotion Challenge and Workshop challenge (AVEC2016) \citep{avec2016} and perform binary classification on the dialogue level. Interviewees with PHQ$8\geq10$ is considered \textit{Depressed} and PHQ$8<10$ is considered \textit{Not Depressed}. Since patients with PHQ-8 score of 5 to 9 are defined to show mild depressive symptoms \citep{Kroenke2008-qm} but considered \textit{Not Depressed} in the dataset, we add information about PHQ-8 level definition and quantisation criteria to the prompt to establish an aligned diagnosis standard (Table \ref{tab:prompt-design}) for the model. 

Notably, participants in the AVEC2016 challenge \citep{daicwoz-baseline-1,daicwoz-baseline-2} and subsequent research efforts \citep{daicwoz-baseline-3,daicwoz-sota} primarily focused on optimising the F1 score of the \textit{Depressed} class. We report this metric in Section \ref{sec:results-and-discussions} for direct comparison and also include the F1 score of the \textit{Not Depressed} in Appendix \ref{sec:detailed-results}.




\subsection{Prompt Design}
\label{sec:prompt-design}
\begin{table*}[h]
    \centering
    \scriptsize
    \setlength\tabcolsep{3pt}
    \begin{tabular}{ll}
    \toprule
         &  Prompt Template \\
    \midrule
    Task & \textbf{IEMOCAP:} Consider the following list of concepts, called \texttt{EMOTION}s: [\texttt{Emotion\textsubscript{A}, Emotion\textsubscript{B},} ...] \\
    Definition & \textbf{EmoWOZ:} Consider the following list of concepts, called \texttt{EMOTION}s: [\texttt{Emotion\textsubscript{A}:} \texttt{Emotion\_Definition\textsubscript{A}}; \texttt{Emotion\textsubscript{B}:} \texttt{Emotion\_Definition\textsubscript{B}}; ...]\\
    & \textbf{DAIC-WOZ:} Given that the \texttt{SEVERITY} of depression can be categorised into the following levels on a scale of 0 to 24: [\texttt{No significant}\\
    & \texttt{depressive symptoms (0 to 4),} ...]. A participant is considered depressed if the participant shows moderate depressive symptoms (10 to 14) \\
    & and above. \\
    \midrule
    ICL & \textbf{IEMOCAP} / \textbf{EmoWOZ:} Given the dialogue history between \texttt{Speaker\textsubscript{A}} and \texttt{Speaker\textsubscript{B}}: [\texttt{Speaker\textsubscript{A}:} \texttt{Utterance\textsubscript{t-3};} \texttt{Speaker\textsubscript{B}:} \texttt{Utterance\textsubscript{t-2};} \\
    Samples & \texttt{Speaker\textsubscript{A}: Utterance\textsubscript{t-1}}], the \verb|EMOTION| in the next utterance ``\texttt{Speaker\textsubscript{B}:} \texttt{Utterance\textsubscript{t}}'' is \texttt{Emotion\textsubscript{A}}\\
    & \textbf{DAIC-WOZ}: Given the depression consultation dialogue between \texttt{Participant} and  \texttt{Ellie}: [\texttt{Participant:} \texttt{Utterance\textsubscript{0}; Ellie:} \texttt{Utterance\textsubscript{1};} \\
    & \texttt{Participant: Utterance\textsubscript{2};} ...], the \texttt{Participant}'s is \texttt{(not) depressed}. \\
    \midrule
    Query & \textbf{IEMOCAP} / \textbf{EmoWOZ:} Given the dialogue history between \texttt{Speaker\textsubscript{A}} and \texttt{Speaker\textsubscript{B}}: [\texttt{Speaker\textsubscript{A}:} \texttt{Utterance\textsubscript{t-3};} \texttt{Speaker\textsubscript{B}:} \texttt{Utterance\textsubscript{t-2};} \\
    & \texttt{Speaker\textsubscript{A}: Utterance\textsubscript{t-1}}], the \verb|EMOTION| in the next utterance ``\texttt{Speaker\textsubscript{B}:} \texttt{Utterance\textsubscript{t}}'' is \\
    & \textbf{DAIC-WOZ}: Given the depression consultation dialogue between \texttt{Participant} and  \texttt{Ellie}: [\texttt{Participant:} \texttt{Utterance\textsubscript{0}; Ellie:} \texttt{Utterance\textsubscript{1};} \\
    & \texttt{Participant: Utterance\textsubscript{2};} ...], the \texttt{Participant}'s is \\
    \bottomrule
    \end{tabular}
\caption{Prompt templates, consisting of the task definition, in-context samples, and the query.\label{tab:prompt-design}}
\end{table*}

The prompt design aims to exploit the language modelling and in-context learning ability of LLMs. Due to the different task set-ups and label sets in each datasets, the prompt templates used are different as illustrated in Table \ref{tab:prompt-design}. Specifically, EmoWOZ uses custom emotion labels, DAIC-WOZ involves mapping from numerical values to binary classes, and IEMOCAP uses generic emotion labels. We therefore provide additional label explanations in the task definition of EmoWOZ and DAIC-WOZ. IEMOCAP on the other hand, contains self-explanatory emotion labels from a generic emotion model and does not come with any special definitions. Therefore, we do not include label definition in the prompt for IEMOCAP. Since IEMOCAP and EmoWOZ involve utterance-level classification whereas DAIC-WOZ involves dialogue-level classification, we used different queries to accommodate this difference. 

\subsection{Models}
\label{sec:models}
\subsubsection{LLMs}
\textbf{GPT-2} \citep{gpt2} has a transformer architecture, pretrained on a substantial English corpus through self-supervised learning. While its size does not make it one of LLMs, it stands as one of the early achievements of OpenAI's GPT models. For our baseline reference, we utilised the version containing 124 million parameters.

\textbf{GPT-3.5}, or ChatGPT \citep{chatgpt}, is a chatbot application developed by OpenAI. It follows a similar architecture as InstructGPT \citep{instructgpt} and was fine-tuned for chat application via reinforcement learning from human feedback (RLHF). It contains 175 billion parameters. Specifically, we used the version released on the $13^{th}$ of June, 2023.

\textbf{GPT-4} \citep{gpt4} is an improved version of GPT-3.5. Its size is six times that of GPT-3.5. Although it is considered a multi-modal model because it additionally accepts images as input, we only explored its text modality. We used the version released on the $13^{th}$ of June, 2023.

\textbf{LLaMA-7B} \citep{llama} is a large and causal language model introduced by Meta AI in 2023. It has transformer decoder architecture, 7 billion parameters and was pre-trained on 1 trillion tokens.

\textbf{Alpaca-7B} \citep{alpaca} is fine-tuned from LLaMA-7B with 52K instruction-following demonstrations generated in the style of self-instruct using \verb|text-davinci-003|, a specific version of InstructGPT \citep{instructgpt}.

\textbf{LLaMA-3-8B} \citep{llama3modelcard} is the most recent model of the LLaMA family, featuring enhanced usefulness and safety. It was pre-trained on 15 trillion tokens. 

\subsubsection{Supervised Models for Comparison}
While comparing zero-shot and few-shot ICL results of LLMs with supervised SOTAs does not paint the fairest picture, it does provide us with insights into how far LLMs are from achieving the performance levels of supervised SOTAs. 

We compare LLMs' performance with the following supervised models on each dataset: \citet{iemocap-sota} for IEMOCAP, \citet{feng-etal-2023-chatter} for EmoWOZ, and \citet{daicwoz-sota} for DAIC-WOZ. Specifically,

\textbf{For IEMOCAP}: \citet{iemocap-sota} proposed an emotion recognition model which takes 1) a time-synchronous representation that fuses the audio features with the corresponding text information at each time step, as well as 2) a time-asynchronous representation that captures the text information embedded across the transcriptions of a number of consecutive utterances. These two types of frame-level vectors, after being pooled in their respective branches with self-attentive layers across the input time window, are fused using an fully connected layer for emotion classification.

\textbf{For EmoWOZ}: \citet{feng-etal-2023-chatter} proposed a model that is dedicated for textual emotion recognition in task-oriented dialogues. Based on a transformer-based classifier that considers the dialogue history and speaker roles, the proposed model adopts data augmentation with chit-chat dialogues, dialogue state features, multi-task classification for emotional aspects, and a distance-based loss that considers the similarity of the custom emotion labels in EmoWOZ.

\textbf{For DAIC-WOZ}: \citet{daicwoz-sota} proposed to extract utterance-level representations from pre-trained speech-based foundation model. The foundation model was further fine-tuned for speech recognition and emotion recognition. The average-pooled dialogue-level features were fed into a depression detection block for binary classification. To address the issue of data sparsity in speech depression detection, authors also performed data augmentation using sub-dialogue shuffling.



\subsection{Training Configurations}
We implement LoRA (Section \ref{sec:task-specific-fine-tuning}) when training LLaMA-7B, Alpaca-7B, and LLaMA-3-8B but not GPT-2. For all open-source LLMs, we constrain the decoding space of the model output to ensure it generates the desired class labels. Details can be found in Appendix \ref{sec:detailed-training-config}.

\subsection{ASR System Specifications}
\label{sec:asr-specifications}


In order to observe how LLMs perform with the presence of substantial ASR errors rather than building a pipeline for speech-based ERC, we use an ``off-the-shelf'' OpenAI Whisper-medium model \citep{whisper}, which has been trained solely on English data and not been fine-tuned. We use a decoding beam size of 3. The text normalisation only involves removing punctuation marks. 
The ASR word error rates (WER) for IEMOCAP and DAIC-WOZ are 12.0\% and 16.5\% respectively.
Since EmoWOZ does not come with raw audio data, we build an ASR simulator. We formulate the simulation as a sequence generation task where the source is the ground-truth text and the target is the ASR-transcribed text (as described in Appendix \ref{sec:asr-simulator}). The resulted simulated WER in EmoWOZ is 17.1\%.




\section{Results and Discussions}
\label{sec:results-and-discussions}

In this section, we aim to answer the questions below. Full results can be found in Appendix \ref{sec:detailed-results}.

\begin{itemize}[noitemsep,topsep=0pt,leftmargin=*]
    \item How do LLMs perform under zero-shot set-up on different types of dialogues? How robust are LLMs to ASR errors?
    \item To what extent can few-shot in-context learning improve LLMs' performance?
    \item For open-source LLMs, can task-specific fine-tuning achieves SOTA performance on each respective dataset?
\end{itemize}

\subsection{Zero-shot Learning}
\label{sec:zero-shot-learning}

\begin{table*}[]
\centering
\small
\begin{tabular}{l|cc|cc|cc|cc}
\toprule
\multirow{2}{*}{Model} & \multicolumn{2}{c|}{IEMOCAP (4-way)} & \multicolumn{2}{c|}{IEMOCAP (5-way)} & \multicolumn{2}{c|}{EmoWOZ} & \multicolumn{2}{c}{DAIC-WOZ } \\
 & \multicolumn{1}{c}{WA ($\uparrow$)} & \multicolumn{1}{c|}{UA ($\uparrow$)} & \multicolumn{1}{c}{WA ($\uparrow$)} & \multicolumn{1}{c|}{UA ($\uparrow$)} & \multicolumn{1}{c}{MF1 ($\uparrow$)} & \multicolumn{1}{c|}{WF1 ($\uparrow$)} & \multicolumn{1}{c}{F1 (dev, $\uparrow$)} & \multicolumn{1}{c}{F1 (test, $\uparrow$)}  \\ \midrule
GPT-2 & 25.8 & 29.2 & 19.0 & 22.3 & 7.3 & 24.0 & 0.0 & 0.0 \\
LLaMA-7B & 41.1 & 40.5 & 35.6 & 33.6 & 1.1 & 0.3 & 47.5 & 52.2 \\
Alpaca-7B & \textbf{48.8} & \textbf{51.4} & \textbf{40.5} & \textbf{36.2} & 24.0 & 44.6 & 47.5 & 53.3 \\
LLaMA-3-8B & 41.8 & 42.5 & 29.4 & 31.7 & 19.7 & 42.4 & 47.1 & 43.2 \\
GPT-3.5 & 42.2 & 37.6 & 37.9 & 35.1 & 39.0 & 40.0 & 54.5 & \textbf{64.3}  \\
GPT-4 &  42.4 & 37.6 & 37.5 & 34.7 & \textbf{52.4} & \textbf{62.3} & \textbf{63.6} & 59.3 \\ \midrule
Supervised SOTA & 77.6 & 78.4 & 73.3 & 74.4 & 65.9 & 83.9 & 88.6 & 85.7  \\ 
\bottomrule
\end{tabular}
\caption{Zero-shot performance of LLMs compared with respective supervised SOTAs. The best zero-shot performance for each metric is made bold. For metrics: WA = weighted average; UA = unweighted average; MF1 = macro-averaged F1 excluding neutral; WF1 = weighted average F1 excluding neutral; F1 = F1 for class \textit{Depressed}.\label{tab:0shot}}
\end{table*}

Table \ref{tab:0shot} summarises LLMs' zero-shot affect recognition performances on the three datasets, and we made the following observations:

\paragraph{LLMs' performance falls short of supervised SOTAs in affect recognition tasks.} Notable gaps are observed when compared the performance achieved by LLMs and supervised SOTAs for all datasets. 

It's noteworthy that although GPT-4, the largest model, underperforms when compared with the supervised SOTA on EmoWOZ, its reported macro-averaged F1 is still comparable to some supervised learning models benchmarked in \citet{feng-etal-2022-emowoz}. This suggests the good capability of GPT-4 in leveraging the label definitions in the prompt to recognise emotions in EmoWOZ, irrespective of their prevalence. Supervised models, however, may be more susceptible to issues such as label imbalance.

\paragraph{Larger models do not necessarily lead to better performance.} 
For IEMOCAP, Alpaca-7B demonstrates the best performance, even surpassing much larger models (GPT-3.5 and GPT-4). Conversely, for EmoWOZ and DAIC-WOZ, the performance generally improves as the model size increased.

While chit-chat utterances in IEMOCAP are labelled with emotion classes from generic emotion models, EmoWOZ's labels are specifically designed to encode the eliciting conditions of emotions in ToDs. This design necessitates more explicit reasoning in ERC within EmoWOZ compared to IEMOCAP. Although LLMs rely on their language modelling capabilities when performing zero-shot ERC, the greater reasoning ability facilitated by the substantial number of parameters in GPT-3.5 and GPT-4 results in improved performance in EmoWOZ.

Likewise in DAIC-WOZ, the recognition is performed for the entire dialogue. Larger models demonstrate greater ability to leverage the more nuanced affective state of the patient in the larger context.

\paragraph{Fine-tuning LLMs with instruction-following demonstrations facilitates more effective utilisation of the prompt.}
In all datasets, Alpaca-7B consistently outperforms LLaMA-7B and even the much more recent LLaMA-3-8B. This indicates that the additional fine-tuning of LLaMA-7B with instruction-following demonstrations has enhanced its capability in ERC. 

LLaMA-7B appears to underperform compared to the much smaller GPT-2 on EmoWOZ. This discrepancy can be explained by LLaMA-7B's strong inclination towards predicting the neutral emotion (F$1=82.1$ with Recall $=100$), which has been excluded from the metric calculation, resulting in the poor reported metrics. Fine-tuning with instruction-following demonstrations, as adopted in Alpaca-7B, effectively leverages the task and label definition in the prompt and reverts this trend. Such an inclination in predicting neutral emotion in LLaMA-7B does not appear in the more recent LLaMA-3-8B. 

\subsection{Zero-shot Learning with Noisy ASR Input}

\begin{table*}[]
\centering
\setlength\tabcolsep{5pt}
\small
\begin{tabular}{l|cc|cc|cc|cc}
\toprule
\multirow{2}{*}{Model} & \multicolumn{2}{c|}{IEMOCAP (4-way)} & \multicolumn{2}{c|}{IEMOCAP (5-way)} & \multicolumn{2}{c|}{EmoWOZ)} & \multicolumn{2}{c}{DAIC-WOZ}  \\
 & \multicolumn{1}{c}{WA ($\uparrow$)} & \multicolumn{1}{c|}{UA ($\uparrow$)} & \multicolumn{1}{c}{WA ($\uparrow$)} & \multicolumn{1}{c|}{UA ($\uparrow$)} &  \multicolumn{1}{c}{MF1 ($\uparrow$)} & \multicolumn{1}{c|}{WF1 ($\uparrow$)} &  \multicolumn{1}{c}{F1 (dev, $\uparrow$)} & \multicolumn{1}{c}{F1 (test, $\uparrow$)}  \\ \midrule
LLaMA-7B & -0.3 & -1.2 & -1.1 & -5.0 & -1.1 & -0.3 & -1.6 & -1.1 \\
Alpaca-7B & -1.3 & -1.8 & -1.8 & -2.6 & +0.3 & -2.0 & -1.6 & +0.0 \\
LLaMA-3-8B & -2.1 & -3.5 & -1.2 & -2.2 & +0.1 & -0.1 & -0.7 & -0.3 \\
GPT-3.5 & +0.1 & -0.1 & +0.2 & 0.0 & +1.2 & -0.2 & -17.0 & -8.3  \\
GPT-4 & -0.5 & -0.5 & -1.1 & -0.7 & +0.9 & -1.5 & -19.2 & -17.6 \\ \midrule
Supervised SOTA & -3.8 & -3.7 & -3.9 & -3.5 & -0.8 & -0.4 & -3.6  & -4.1  \\
\bottomrule
\end{tabular}
\caption{Change in zero-shot performance metrics of LLMs after using noisy ASR input. For metrics: WA = weighted average; UA = unweighted average; F1 = F1 for class \textit{Depressed}. GPT-2 was omitted due to its poor zero-shot capability.\label{tab:0shot-asr}}
\end{table*}

Table \ref{tab:0shot-asr} provides a summary of LLMs' zero-shot performance when replacing the original dialogue transcripts with ASR-inferred inputs. ASR errors exhibit varying degrees of influence on different affect recognition tasks. Specifically,

\paragraph{LLMs are generally robust to ASR errors when recognising emotion.} This is exemplified by small changes in metrics for IEMOCAP compared with supervised SOTAs. The only one notable exception is the UA of LLaMA-7B in the 5-way classification task on IEMOCAP. Looking at the performance of each emotion in this experiment, we observed significant drops in the F1 scores for the emotions \{\textit{Happy}, \textit{Angry}, and \textit{Sad}\}. Specifically, \textit{Happy} and \textit{Angry} experience major decreases in their recall scores (\textit{Happy}: $12.3 \rightarrow 7.3$, \textit{Angry}: $50.0 \rightarrow 11.0$), while \textit{Sad} sees a substantial decline in its precision score ($65.5 \rightarrow 0.0$). At the same time, there is an increase in the recall score for the \textit{Other} category ($47.3 \rightarrow 78.2$), resulting in an overall rise in its F1 score ($44.5 \rightarrow 48.0$). These observations suggest that ASR errors introduced a tendency for LLaMA-7B to mis-classify more emotions as \textit{Other}.

\paragraph{ASR errors have a more pronounced influence on the accuracy of depression detection.} For DAIC-WOZ, the introduction of ASR errors had a significant impact on F1 scores. The impact diverges for open-source and commercial models. For open-source models, which are also relatively smaller, the change in F1 was small, showing a similar trend when they recognise emotions from noisy dialogues. On the other hand, for larger commercial models, the F1 scores decrease more significantly.
This phenomenon can be ascribed to the lengthy prompt for conducting dialogue-level analysis, in which ASR errors accumulated. While OpenAI models can better leverage information from the large context, such an ability adversely affects its depression detection ability in the presence of ASR errors.

\subsection{In-context Learning}

\begin{table}[h!]
\centering
\scriptsize
\begin{tabular}{lcccccc}
\toprule
\multirow{2}{*}{Model} & \multirow{2}{*}{N} & \multicolumn{2}{c}{IEMOCAP} & \multirow{2}{*}{EmoWOZ} & \multicolumn{2}{c}{DAIC-WOZ} \\
 &  & 4-way & 5-way & &  Dev & Test \\ \midrule
 & 0 & 41.1 & \textbf{35.6} & 0.3  & \textbf{47.5} & \textbf{52.2} \\
LLaMA-7B & 1 & \textbf{52.3} & 27.3 & \textbf{42.6} & 0.0 & 0.0 \\
 & 3 & 42.8 & 26.2 & 27.2 & 42.1 & 48.9 \\ \midrule
 & 0 & 48.8 & \textbf{40.5} & 44.6 & \textbf{47.5} & \textbf{53.3} \\ 
Alpaca-7B & 1 & \textbf{54.1} & 26.9 & \textbf{51.2} & 0.0 & 15.4 \\
 & 3 & 52.4 & 24.4 & 44.6 & 45.9 & 51.1 \\ \midrule
 & 0 & 41.8 & 29.4 & \textbf{42.4} & \textbf{47.1} & \textbf{43.2} \\
LLaMA-3-8B & 1 & 56.8 & \textbf{40.5} & 38.0 & 0.0 & 0.0 \\
 & 3 & \textbf{57.4} & 24.4 & 39.9 & 0.0 & 0.0 \\ \midrule
 & 0 & 42.2 & 37.9 & 40.0 & \textbf{54.5} & \textbf{64.3} \\
GPT-3.5 & 1 & 56.3 & \textbf{48.3} & 43.2 & 13.3 & 40.0 \\
 & 3 & \textbf{62.1} & \textbf{48.3} & \textbf{46.7} & 37.5 & 56.0 \\ \midrule
 & 0 & 42.4 & 37.5 & 62.3 & 63.6 & \textbf{59.3} \\
GPT-4 & 1 & 62.9 & 49.0 & 64.4 & \textbf{80.0} & 55.6 \\ 
 & 3 & \textbf{63.8} & \textbf{49.4} & \textbf{66.5} & 74.1 & 58.5 \\ 
\bottomrule
\end{tabular} 
\caption{Performance of LLMs (WA for IEMOCAP and WF1 for EmoWOZ) under in-context learning set-ups. N stands for the number of ICL samples per emotion class and $N=0$ means the zero-shot set-up. The best performance of each model is made bold. \label{tab:icl-exp}}
\end{table}

\begin{figure*}[h]
\centering
 \includegraphics[width=0.9\textwidth]{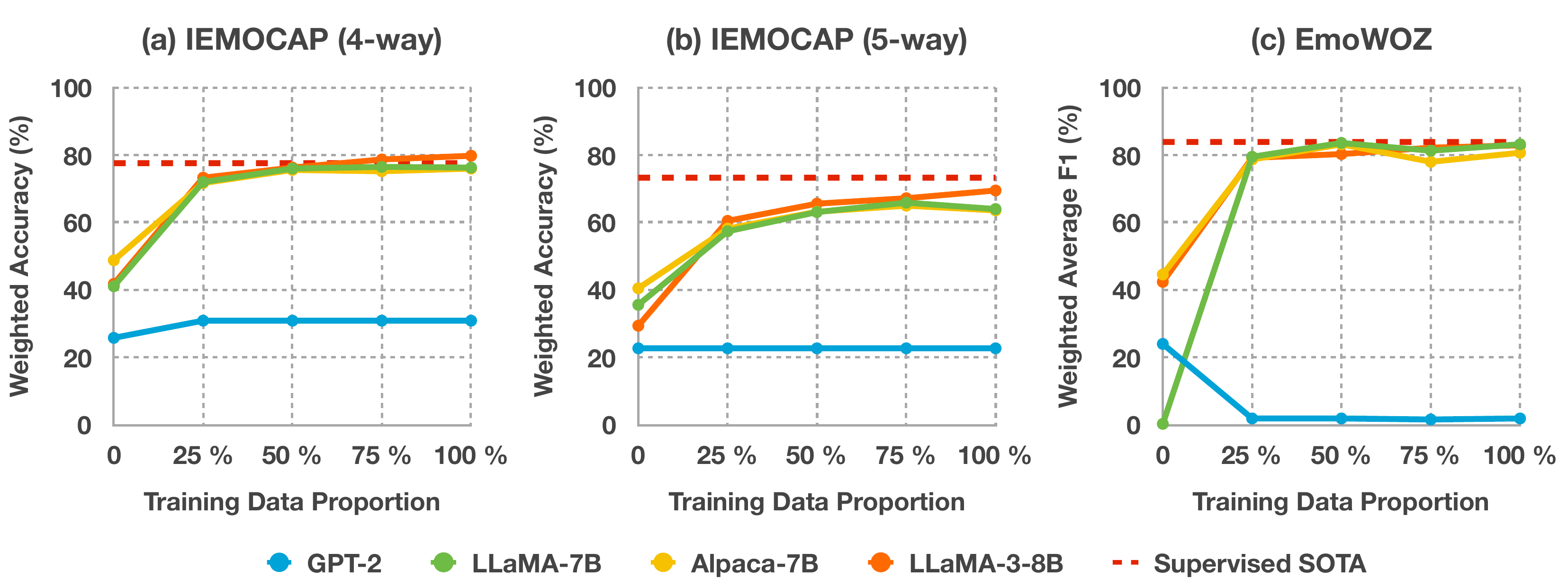}
 \caption{Change of model performance when fine-tuning with different proportions of the training data. \label{fig:ft_proportions}}
\end{figure*}

ICL samples are randomly selected for each class and are the same within each experiment set-up for all models. The performance of LLMs with different numbers of ICL samples is outlined in Table \ref{tab:icl-exp}, from which we have derived the following observation:

\paragraph{Larger models tend to derive greater benefits from an increased number of ICL samples to recognise emotions.}
LLaMA-7B, Alpaca-7B, and LLaMA-3-8B do not consistently benefit from an increased number of ICL samples in the prompt. Optimal model performance generally occurs when $N=0$ or $N=1$. 
This suggests that effectively utilising the full context remains as a challenge for LLMs.
Larger models, GPT-3.5 and GPT-4, show more consistent improvement in performance with the increased number of ICL samples. GPT-4 derives the most significant benefits from ICL samples and performs the best across all models. 

\paragraph{The effectiveness of ICL is limited for depression detection.} The performance is in general the best when $N=0$, followed by $N=3$. This suggests that for depression detection, a task to detect more nuanced affective state than emotion from a longer sequence, a single ICL sample for each class could strongly bias the model. This leads to zero F1s where models predict all samples as \textit{Not Depressive}. Including more ICL samples could mitigate this effect, but the performance is further limited by models' incapability to handle extremely lengthy input. This motivates further research efforts to handle huge context containing nuanced task-related cues when using LLMs.





\subsection{Task-specific Fine-tuning}

We conduct task-specific fine-tuning experiments with GPT-2, LLaMA-7B, Alpaca-7B, and LLaMA-3-8B using different proportions of training data to explore these models' capacity for ERC after fine-tuning. Results are summarised in Figure \ref{fig:ft_proportions}. For DAIC-WOZ, fine-tuning would steer models to predict \textit{Not Depressed} (see Table \ref{tab:daicwoz-full}) for almost all test samples. This might be due to the small training set where more than 70\% of the samples are labelled as \textit{Not Depressed}. This suggests the limitation of language modelling objective, and therefore more carefully curated task-related learning objectives should be considered for depression detection using LLM.

\paragraph{Task-specific fine-tuning can effectively and efficiently enhance the ERC performance of LLMs.}
For both IEMOCAP and EmoWOZ, we observe an initial significant improvement in performance when fine-tuning with 25\% of the training data. Performance remains relatively stable and approaches SOTA levels as the proportion of training data increased to 50\% and more for IEMOCAP (4-way) and EmoWOZ. This shows the potential of rapid deployment of LLMs as the emotion recognition frontend in dialogue systems, regardless of dialogue type, label set, or label distribution. 

In the case of 5-way classification on IEMOCAP, a performance gap persists between fine-tuned LLMs and the supervised SOTA, even after fine-tuning of LLMs on the complete training set. We hypothesised that this disparity might be attributed to the presence of an additional \textit{Other} class within the 5-way classification scheme. The class name ``Other'' lacked essential affective information and consequently failed to fully leverage the language modelling capabilities of LLMs. Therefore, we suggest that employing more semantically meaningful label names could be advantageous in harnessing the potential of LLMs for task-specific fine-tuning.

In the case of GPT-2, fine-tuning does not yield noticeable improvement in ERC. Its performance even deteriorated after fine-tuning with EmoWOZ, as depicted in Figure \ref{fig:ft_proportions}(c) because GPT-2 predominantly predicted \textit{Neutral}, which are excluded from the metric calculation.

\section{Conclusion}
In this study, we explore the performance of LLMs for affect recognition in three distinct types of dialogues: chit-chat dialogues, information-seeking ToDs, and medical consultation dialogues for depression. We conduct benchmark experiments on these datasets using five LLMs: LLaMA-7B, Alpaca-7B, LLaMA-3-8B, GPT-3.5, and GPT-4. We also explore various setups, including zero-shot learning, few-shot in-context learning, and task-specific fine-tuning, all facilitated by specially designed prompts. Additionally, we examine the impact of ASR errors on LLMs' zero-shot performance.

Our zero-shot experiments underscore that while LLMs have made significant strides in various natural language understanding tasks, they still have some distance to cover in order to match the supervised SOTAs in affect recognition tasks. Adding emotion definitions explaining the eliciting conditions in ToDs to the prompt and fine-tuning LLMs for instruction-following could narrow the performance gap from supervised SOTAs.

Performing zero-shot affect recognition from utterances containing ASR errors shows that LLMs are robust to such errors for emotion recognition but not for depression detection. Therefore, when considering LLMs as a back-end module of a spoken dialogue system, it is crucial to exercise extra caution when processing dialogues laden with highly specific and nuanced affective content.

Our ICL experiments exemplify that larger models would benefit more from an increased number of ICL samples, highlighting the need to explore the optimal combination of the ICL sample size in the prompt and the model size. For smaller LLMs, effectively utilising lengthy context remains as a challenge.

Through task-specific fine-tuning, we achieve performance levels close to SOTA on IEMOCAP and EmoWOZ, using only 50\% of the training data, with LLaMA-7B, Alpaca-7B, and LLaMA-3-8B. This highlights the great potential of fine-tuning LLMs for simpler tasks and integrating them as functional modules into dialogue systems.

Overall, LLMs have opened new avenues for affect recognition in conversations and building affect-aware dialogue systems. Despite the limited performance under zero-shot set-up, their robustness to ASR errors, few-shot ICL capabilities, and ERC capabilities after fine-tuning offer exciting research opportunities for exploring affect recognition in conversations and building human-like conversational agents. We would also like to highlight the challenge and also opportunities towards handling long context and nuanced emotion cues in LLMs.

\section{Limitations}
In our work, although we reduce computation resource of training LLMs by incorporating LoRA, the inference takes \~1s for utterance-level emotion recognition on a Nvidia A100 40GB graphics card when there is no ICL sample in the prompt. The inference time increases when the number of ICL samples increases or dialogue-level classification is performed. While LLMs demonstrates superior abilities and potentials, further research efforts are still needed to ensure efficient LLM inference, which is necessary for its application in real-time systems. 

With ICL experiments especially on DAIC-WOZ, we observe that the efficacy of long context is limited by the effective spans of the attention mechanisms. While substantial efforts have been invested into increasing the maximum allowed context size of LLMs and improving benchmark performance, the effectiveness of LLMs to make use of full context should not be overlooked.

We only investigate with one dataset from each of three dialogue domains. Although these datasets cover different dialogue settings, objectives, label sets, and classification scopes, there are more affect types and dialogue settings to explore. These datasets also exhibit various degrees of class imbalance, which selected reference SOTAs utilised data augmentation to address. While GPT-4 has demonstrated good zero-shot learning ability (Section \ref{sec:zero-shot-learning}), addressing data imbalance is out of the scope of this work, and data augmentation with LLMs may come at a cost of potential divergence between synthetic language and real-word data \citep{li-etal-2023-synthetic}.

\section{Ethics Statement}
Models and datasets were used in accordance with their respective licenses. Data that we used and generated does not contain any information that uniquely identifies individual people. There is a tiny fraction of utterances labelled as ``abusive'' in EmoWOZ, but they are prompted to models in such a way for the recognition purpose only. Due to the fact that LLMs were pre-trained with a huge amount of data, they may produce inaccurate information about people, places, or facts. This had negligible impact on our evaluation for affect recognition. When performing depression detection and analysis with DAIC-WOZ using GPT-3.5 and GPT-4, models output reminders about seeking professional advice from doctors for more accurate medical diagnosis along with their predictions. 

Unlike running models locally, utilising OpenAI's server-based models would require us to send data to their server. In some cases, it is important to use the application programming interface (API) when for which OpenAI explicitly clarifies that the query data will not be stored or used in model training unless specifically configured.

Although this work focuses on LLMs' capability in recognising affect in conversations, we do envisage LLMs to be incorporated as an affect recognition frontend in affect-aware dialogue systems. It is therefore important to remember that these models are not perfect and can make errors in their predictions. Subsequently, any actions taken based on these predictions should be executed with an awareness of the possibility of errors. The relatively slow inference speed and the high computational resource requirement also pose a challenge in the usage of LLMs in high-throughput and time-sensitive scenarios.

\section{Acknowledgement}
S. Feng and N. Lubis are supported by funding provided by the Alexander von Humboldt Foundation in the framework of the Sofja Kovalevskaja Award endowed by the Federal Ministry of Education and Research. G. Sun is partly funded by the Department of Engineering, University of Cambridge. Computing resources were provided by Google Cloud.

\bibliography{anthology,custom}

\begin{thebibliography}{42}
\expandafter\ifx\csname natexlab\endcsname\relax\def\natexlab#1{#1}\fi

\bibitem[{AI@Meta(2024)}]{llama3modelcard}
AI@Meta. 2024.
\newblock \href {https://github.com/meta-llama/llama3/blob/main/MODEL_CARD.md} {Llama 3 model card}.

\bibitem[{{American Psychiatric Association}(2020)}]{depression-definition}
{American Psychiatric Association}. 2020.
\newblock {What Is Depression?}
\newblock \url{https://www.psychiatry.org/patients-families/depression/what-is-depression}.

\bibitem[{Beeching et~al.(2023)Beeching, Fourrier, Habib, Han, Lambert, Rajani, Sanseviero, Tunstall, and Wolf}]{open-llm-leaderboard}
Edward Beeching, Clémentine Fourrier, Nathan Habib, Sheon Han, Nathan Lambert, Nazneen Rajani, Omar Sanseviero, Lewis Tunstall, and Thomas Wolf. 2023.
\newblock Open llm leaderboard.
\newblock \url{https://huggingface.co/spaces/HuggingFaceH4/open_llm_leaderboard}.

\bibitem[{Budzianowski et~al.(2018)Budzianowski, Wen, Tseng, Casanueva, Ultes, Ramadan, and Ga{\v{s}}i{\'c}}]{budzianowski-etal-2018-multiwoz}
Pawe{\l} Budzianowski, Tsung-Hsien Wen, Bo-Hsiang Tseng, I{\~n}igo Casanueva, Stefan Ultes, Osman Ramadan, and Milica Ga{\v{s}}i{\'c}. 2018.
\newblock \href {https://doi.org/10.18653/v1/D18-1547} {{M}ulti{WOZ} - a large-scale multi-domain {W}izard-of-{O}z dataset for task-oriented dialogue modelling}.
\newblock In \emph{Proceedings of the 2018 Conference on Empirical Methods in Natural Language Processing}, pages 5016--5026, Brussels, Belgium. Association for Computational Linguistics.

\bibitem[{Busso et~al.(2008)Busso, Bulut, Lee, Kazemzadeh, Mower, Kim, Chang, Lee, and Narayanan}]{iemocap}
Carlos Busso, Murtaza Bulut, Chi-Chun Lee, Abe Kazemzadeh, Emily Mower, Samuel Kim, Jeannette~N. Chang, Sungbok Lee, and Shrikanth~S. Narayanan. 2008.
\newblock {IEMOCAP}: interactive emotional dyadic motion capture database.
\newblock \emph{Language Resources and Evaluation}, 42(4):335--359.

\bibitem[{Ekman and Friesen(1971)}]{ekman-basic}
Paul Ekman and W~V Friesen. 1971.
\newblock \href {https://api.semanticscholar.org/CorpusID:14013552} {Constants across cultures in the face and emotion.}
\newblock \emph{Journal of personality and social psychology}, 17 2:124--9.

\bibitem[{Feng et~al.(2022)Feng, Lubis, Geishauser, Lin, Heck, van Niekerk, and Gasic}]{feng-etal-2022-emowoz}
Shutong Feng, Nurul Lubis, Christian Geishauser, Hsien-chin Lin, Michael Heck, Carel van Niekerk, and Milica Gasic. 2022.
\newblock \href {https://aclanthology.org/2022.lrec-1.436} {{E}mo{WOZ}: A large-scale corpus and labelling scheme for emotion recognition in task-oriented dialogue systems}.
\newblock In \emph{Proceedings of the Thirteenth Language Resources and Evaluation Conference}, pages 4096--4113, Marseille, France. European Language Resources Association.

\bibitem[{Feng et~al.(2023)Feng, Lubis, Ruppik, Geishauser, Heck, Lin, van Niekerk, Vukovic, and Gasic}]{feng-etal-2023-chatter}
Shutong Feng, Nurul Lubis, Benjamin Ruppik, Christian Geishauser, Michael Heck, Hsien-chin Lin, Carel van Niekerk, Renato Vukovic, and Milica Gasic. 2023.
\newblock \href {https://aclanthology.org/2023.sigdial-1.8} {From chatter to matter: Addressing critical steps of emotion recognition learning in task-oriented dialogue}.
\newblock In \emph{Proceedings of the 24th Meeting of the Special Interest Group on Discourse and Dialogue}, pages 85--103, Prague, Czechia. Association for Computational Linguistics.

\bibitem[{Gratch et~al.(2014)Gratch, Artstein, Lucas, Stratou, Scherer, Nazarian, Wood, Boberg, DeVault, Marsella, Traum, Rizzo, and Morency}]{gratch-etal-2014-distress}
Jonathan Gratch, Ron Artstein, Gale Lucas, Giota Stratou, Stefan Scherer, Angela Nazarian, Rachel Wood, Jill Boberg, David DeVault, Stacy Marsella, David Traum, Skip Rizzo, and Louis-Philippe Morency. 2014.
\newblock \href {http://www.lrec-conf.org/proceedings/lrec2014/pdf/508_Paper.pdf} {The distress analysis interview corpus of human and computer interviews}.
\newblock In \emph{Proceedings of the Ninth International Conference on Language Resources and Evaluation ({LREC}'14)}, pages 3123--3128, Reykjavik, Iceland. European Language Resources Association (ELRA).

\bibitem[{Gross(2002)}]{gross_2002}
James~J. Gross. 2002.
\newblock \href {https://doi.org/10.1017/S0048577201393198} {Emotion regulation: Affective, cognitive, and social consequences}.
\newblock \emph{Psychophysiology}, 39(3):281–291.

\bibitem[{Heck et~al.(2023)Heck, Lubis, Ruppik, Vukovic, Feng, Geishauser, Lin, van Niekerk, and Gasic}]{chatgpt-dst}
Michael Heck, Nurul Lubis, Benjamin Ruppik, Renato Vukovic, Shutong Feng, Christian Geishauser, Hsien-chin Lin, Carel van Niekerk, and Milica Gasic. 2023.
\newblock \href {https://doi.org/10.18653/v1/2023.acl-short.81} {{C}hat{GPT} for zero-shot dialogue state tracking: A solution or an opportunity?}
\newblock In \emph{Proceedings of the 61st Annual Meeting of the Association for Computational Linguistics (Volume 2: Short Papers)}, pages 936--950, Toronto, Canada. Association for Computational Linguistics.

\bibitem[{Hu et~al.(2022)Hu, Shen, Wallis, Allen-Zhu, Li, Wang, Wang, and Chen}]{lora}
Edward~J Hu, Yelong Shen, Phillip Wallis, Zeyuan Allen-Zhu, Yuanzhi Li, Shean Wang, Lu~Wang, and Weizhu Chen. 2022.
\newblock \href {https://openreview.net/forum?id=nZeVKeeFYf9} {Lo{RA}: Low-rank adaptation of large language models}.
\newblock In \emph{International Conference on Learning Representations}.

\bibitem[{Huang et~al.(2023)Huang, Lam, Li, Ren, Wang, Jiao, Tu, and Lyu}]{llm-empathy}
Jen-tse Huang, Man Lam, Eric Li, Shujie Ren, Wenxuan Wang, Wenxiang Jiao, Zhaopeng Tu, and Michael Lyu. 2023.
\newblock \href {http://arxiv.org/abs/2308.03656} {Emotionally numb or empathetic? evaluating how llms feel using emotionbench}.

\bibitem[{Jurafsky and Martin(2009)}]{jurafsky}
Dan Jurafsky and James~H. Martin. 2009.
\newblock \href {http://www.amazon.com/Speech-Language-Processing-2nd-Edition/dp/0131873210/ref=pd_bxgy_b_img_y} {\emph{Speech and language processing : an introduction to natural language processing, computational linguistics, and speech recognition}}.
\newblock Pearson Prentice Hall, Upper Saddle River, N.J.

\bibitem[{Kroenke et~al.(2008)Kroenke, Strine, Spitzer, Williams, Berry, and Mokdad}]{Kroenke2008-qm}
Kurt Kroenke, Tara~W Strine, Robert~L Spitzer, Janet B~W Williams, Joyce~T Berry, and Ali~H Mokdad. 2008.
\newblock The {PHQ-8} as a measure of current depression in the general population.
\newblock \emph{J Affect Disord}, 114(1-3):163--173.

\bibitem[{Li et~al.(2023)Li, Zhu, Lu, and Yin}]{li-etal-2023-synthetic}
Zhuoyan Li, Hangxiao Zhu, Zhuoran Lu, and Ming Yin. 2023.
\newblock \href {https://doi.org/10.18653/v1/2023.emnlp-main.647} {Synthetic data generation with large language models for text classification: Potential and limitations}.
\newblock In \emph{Proceedings of the 2023 Conference on Empirical Methods in Natural Language Processing}, pages 10443--10461, Singapore. Association for Computational Linguistics.

\bibitem[{Mangrulkar et~al.(2022)Mangrulkar, Gugger, Debut, Belkada, Paul, and Bossan}]{peft}
Sourab Mangrulkar, Sylvain Gugger, Lysandre Debut, Younes Belkada, Sayak Paul, and Benjamin Bossan. 2022.
\newblock Peft: State-of-the-art parameter-efficient fine-tuning methods.
\newblock \url{https://github.com/huggingface/peft}.

\bibitem[{Mayer et~al.(1999)Mayer, Caruso, and Salovey}]{MAYER1999267}
John~D Mayer, David~R Caruso, and Peter Salovey. 1999.
\newblock \href {https://doi.org/https://doi.org/10.1016/S0160-2896(99)00016-1} {Emotional intelligence meets traditional standards for an intelligence}.
\newblock \emph{Intelligence}, 27(4):267--298.

\bibitem[{{OpenAI}(2022)}]{chatgpt}
{OpenAI}. 2022.
\newblock Introducing {ChatGPT}.
\newblock \url{https://openai.com/blog/chatgpt}.

\bibitem[{OpenAI(2023)}]{gpt4}
OpenAI. 2023.
\newblock \href {http://arxiv.org/abs/2303.08774} {Gpt-4 technical report}.

\bibitem[{Ouyang et~al.(2022)Ouyang, Wu, Jiang, Almeida, Wainwright, Mishkin, Zhang, Agarwal, Slama, Ray, Schulman, Hilton, Kelton, Miller, Simens, Askell, Welinder, Christiano, Leike, and Lowe}]{instructgpt}
Long Ouyang, Jeff Wu, Xu~Jiang, Diogo Almeida, Carroll~L. Wainwright, Pamela Mishkin, Chong Zhang, Sandhini Agarwal, Katarina Slama, Alex Ray, John Schulman, Jacob Hilton, Fraser Kelton, Luke Miller, Maddie Simens, Amanda Askell, Peter Welinder, Paul Christiano, Jan Leike, and Ryan Lowe. 2022.
\newblock \href {https://doi.org/https://doi.org/10.48550/arXiv.2203.02155} {Training language models to follow instructions with human feedback}.
\newblock In \emph{Proceedings of NeurIPS 2022}.

\bibitem[{Picard(1997)}]{Picard97}
Rosalind~W. Picard. 1997.
\newblock \emph{Affective Computing}.
\newblock MIT Press, Cambridge, MA.

\bibitem[{Radford et~al.(2022)Radford, Kim, Xu, Brockman, McLeavey, and Sutskever}]{whisper}
Alec Radford, Jong~Wook Kim, Tao Xu, Greg Brockman, Christine McLeavey, and Ilya Sutskever. 2022.
\newblock \href {http://arxiv.org/abs/2212.04356} {Robust speech recognition via large-scale weak supervision}.

\bibitem[{Radford et~al.(2019)Radford, Wu, Child, Luan, Amodei, and Sutskever}]{gpt2}
Alec Radford, Jeff Wu, Rewon Child, David Luan, Dario Amodei, and Ilya Sutskever. 2019.
\newblock Language models are unsupervised multitask learners.

\bibitem[{Ravi et~al.(2022)Ravi, Wang, Flint, and Alwan}]{daicwoz-baseline-3}
Vijay Ravi, Jinhan Wang, Jonathan Flint, and Abeer Alwan. 2022.
\newblock A step towards preserving speakers' identity while detecting depression via speaker disentanglement.
\newblock \emph{Interspeech}, 2022:3338--3342.

\bibitem[{Russell(1980)}]{1981-25062-001}
James~A. Russell. 1980.
\newblock \href {https://doi.org/10.1037/h0077714} {A circumplex model of affect.}
\newblock \emph{Journal of Personality and Social Psychology}, 39(6):1161--1178.

\bibitem[{Sun et~al.(2023)Sun, Feng, Jiang, Zhang, Gašić, and Woodland}]{sun2023speechbasedslotfillingusing}
Guangzhi Sun, Shutong Feng, Dongcheng Jiang, Chao Zhang, Milica Gašić, and Philip~C. Woodland. 2023.
\newblock \href {http://arxiv.org/abs/2311.07418} {Speech-based slot filling using large language models}.

\bibitem[{Tang et~al.(2024)Tang, Yu, Sun, Chen, Tan, Li, Lu, MA, and Zhang}]{tang2024salmonn}
Changli Tang, Wenyi Yu, Guangzhi Sun, Xianzhao Chen, Tian Tan, Wei Li, Lu~Lu, Zejun MA, and Chao Zhang. 2024.
\newblock \href {https://openreview.net/forum?id=14rn7HpKVk} {{SALMONN}: Towards generic hearing abilities for large language models}.
\newblock In \emph{The Twelfth International Conference on Learning Representations}.

\bibitem[{Taori et~al.(2023)Taori, Gulrajani, Zhang, Dubois, Li, Guestrin, Liang, and Hashimoto}]{alpaca}
Rohan Taori, Ishaan Gulrajani, Tianyi Zhang, Yann Dubois, Xuechen Li, Carlos Guestrin, Percy Liang, and Tatsunori~B. Hashimoto. 2023.
\newblock Stanford alpaca: An instruction-following llama model.
\newblock \url{https://github.com/tatsu-lab/stanford_alpaca}.

\bibitem[{Touvron et~al.(2023{\natexlab{a}})Touvron, Lavril, Izacard, Martinet, Lachaux, Lacroix, Rozière, Goyal, Hambro, Azhar, Rodriguez, Joulin, Grave, and Lample}]{llama}
Hugo Touvron, Thibaut Lavril, Gautier Izacard, Xavier Martinet, Marie-Anne Lachaux, Timothée Lacroix, Baptiste Rozière, Naman Goyal, Eric Hambro, Faisal Azhar, Aurelien Rodriguez, Armand Joulin, Edouard Grave, and Guillaume Lample. 2023{\natexlab{a}}.
\newblock \href {http://arxiv.org/abs/2302.13971} {Llama: Open and efficient foundation language models}.

\bibitem[{Touvron et~al.(2023{\natexlab{b}})Touvron, Martin, Stone, Albert, Almahairi, Babaei, Bashlykov, Batra, Bhargava, Bhosale, Bikel, Blecher, Ferrer, Chen, Cucurull, Esiobu, Fernandes, Fu, Fu, Fuller, Gao, Goswami, Goyal, Hartshorn, Hosseini, Hou, Inan, Kardas, Kerkez, Khabsa, Kloumann, Korenev, Koura, Lachaux, Lavril, Lee, Liskovich, Lu, Mao, Martinet, Mihaylov, Mishra, Molybog, Nie, Poulton, Reizenstein, Rungta, Saladi, Schelten, Silva, Smith, Subramanian, Tan, Tang, Taylor, Williams, Kuan, Xu, Yan, Zarov, Zhang, Fan, Kambadur, Narang, Rodriguez, Stojnic, Edunov, and Scialom}]{llama2}
Hugo Touvron, Louis Martin, Kevin Stone, Peter Albert, Amjad Almahairi, Yasmine Babaei, Nikolay Bashlykov, Soumya Batra, Prajjwal Bhargava, Shruti Bhosale, Dan Bikel, Lukas Blecher, Cristian~Canton Ferrer, Moya Chen, Guillem Cucurull, David Esiobu, Jude Fernandes, Jeremy Fu, Wenyin Fu, Brian Fuller, Cynthia Gao, Vedanuj Goswami, Naman Goyal, Anthony Hartshorn, Saghar Hosseini, Rui Hou, Hakan Inan, Marcin Kardas, Viktor Kerkez, Madian Khabsa, Isabel Kloumann, Artem Korenev, Punit~Singh Koura, Marie-Anne Lachaux, Thibaut Lavril, Jenya Lee, Diana Liskovich, Yinghai Lu, Yuning Mao, Xavier Martinet, Todor Mihaylov, Pushkar Mishra, Igor Molybog, Yixin Nie, Andrew Poulton, Jeremy Reizenstein, Rashi Rungta, Kalyan Saladi, Alan Schelten, Ruan Silva, Eric~Michael Smith, Ranjan Subramanian, Xiaoqing~Ellen Tan, Binh Tang, Ross Taylor, Adina Williams, Jian~Xiang Kuan, Puxin Xu, Zheng Yan, Iliyan Zarov, Yuchen Zhang, Angela Fan, Melanie Kambadur, Sharan Narang, Aurelien Rodriguez, Robert Stojnic, Sergey Edunov, and Thomas
  Scialom. 2023{\natexlab{b}}.
\newblock \href {http://arxiv.org/abs/2307.09288} {Llama 2: Open foundation and fine-tuned chat models}.

\bibitem[{Valstar et~al.(2016)Valstar, Gratch, Schuller, Ringeval, Lalanne, Torres~Torres, Scherer, Stratou, Cowie, and Pantic}]{avec2016}
Michel Valstar, Jonathan Gratch, Bj\"{o}rn Schuller, Fabien Ringeval, Denis Lalanne, Mercedes Torres~Torres, Stefan Scherer, Giota Stratou, Roddy Cowie, and Maja Pantic. 2016.
\newblock \href {https://doi.org/10.1145/2988257.2988258} {Avec 2016: Depression, mood, and emotion recognition workshop and challenge}.
\newblock In \emph{Proceedings of the 6th International Workshop on Audio/Visual Emotion Challenge}, AVEC '16, page 3–10, New York, NY, USA. Association for Computing Machinery.

\bibitem[{Wang et~al.(2023)Wang, Li, Yin, Wu, and Jia}]{llm-eq}
Xuena Wang, Xueting Li, Zi~Yin, Yue Wu, and Liu Jia. 2023.
\newblock \href {http://arxiv.org/abs/2307.09042} {Emotional intelligence of large language models}.

\bibitem[{Wei et~al.(2022)Wei, Tay, Bommasani, Raffel, Zoph, Borgeaud, Yogatama, Bosma, Zhou, Metzler, Chi, Hashimoto, Vinyals, Liang, Dean, and Fedus}]{emergent}
Jason Wei, Yi~Tay, Rishi Bommasani, Colin Raffel, Barret Zoph, Sebastian Borgeaud, Dani Yogatama, Maarten Bosma, Denny Zhou, Donald Metzler, Ed~H. Chi, Tatsunori Hashimoto, Oriol Vinyals, Percy Liang, Jeff Dean, and William Fedus. 2022.
\newblock \href {https://openreview.net/forum?id=yzkSU5zdwD} {Emergent abilities of large language models}.
\newblock \emph{Transactions on Machine Learning Research}.
\newblock Survey Certification.

\bibitem[{Williamson et~al.(2016)Williamson, Godoy, Cha, Schwarzentruber, Khorrami, Gwon, Kung, Dagli, and Quatieri}]{daicwoz-baseline-2}
James~R. Williamson, Elizabeth Godoy, Miriam Cha, Adrianne Schwarzentruber, Pooya Khorrami, Youngjune Gwon, Hsiang-Tsung Kung, Charlie Dagli, and Thomas~F. Quatieri. 2016.
\newblock \href {https://doi.org/10.1145/2988257.2988263} {Detecting depression using vocal, facial and semantic communication cues}.
\newblock In \emph{Proceedings of the 6th International Workshop on Audio/Visual Emotion Challenge}, AVEC '16, page 11–18, New York, NY, USA. Association for Computing Machinery.

\bibitem[{Wu et~al.(2020)Wu, Zhang, and Woodland}]{iemocap-sota}
Wen Wu, Chao Zhang, and Philip~C. Woodland. 2020.
\newblock \href {https://api.semanticscholar.org/CorpusID:225076233} {Emotion recognition by fusing time synchronous and time asynchronous representations}.
\newblock \emph{ICASSP 2021 - 2021 IEEE International Conference on Acoustics, Speech and Signal Processing (ICASSP)}, pages 6269--6273.

\bibitem[{Wu et~al.(2023)Wu, Zhang, and Woodland}]{daicwoz-sota}
Wen Wu, Chao Zhang, and Philip~C. Woodland. 2023.
\newblock \href {https://doi.org/10.1109/icassp49357.2023.10094910} {Self-supervised representations in speech-based depression detection}.
\newblock In \emph{{ICASSP} 2023 - 2023 {IEEE} International Conference on Acoustics, Speech and Signal Processing ({ICASSP})}. {IEEE}.

\bibitem[{Yang et~al.(2016)Yang, Jiang, He, Pei, Oveneke, and Sahli}]{daicwoz-baseline-1}
Le~Yang, Dongmei Jiang, Lang He, Ercheng Pei, Meshia~C\'{e}dric Oveneke, and Hichem Sahli. 2016.
\newblock \href {https://doi.org/10.1145/2988257.2988269} {Decision tree based depression classification from audio video and language information}.
\newblock In \emph{Proceedings of the 6th International Workshop on Audio/Visual Emotion Challenge}, AVEC '16, page 89–96, New York, NY, USA. Association for Computing Machinery.

\bibitem[{Zeng et~al.(2009)Zeng, Pantic, Roisman, and Huang}]{4468714}
Zhihong Zeng, Maja Pantic, Glenn~I. Roisman, and Thomas~S. Huang. 2009.
\newblock \href {https://doi.org/10.1109/TPAMI.2008.52} {A survey of affect recognition methods: Audio, visual, and spontaneous expressions}.
\newblock \emph{IEEE Transactions on Pattern Analysis and Machine Intelligence}, 31(1):39--58.

\bibitem[{Zhang et~al.(2023)Zhang, Deng, Liu, Pan, and Bing}]{zhang2023sentiment}
Wenxuan Zhang, Yue Deng, Bing Liu, Sinno~Jialin Pan, and Lidong Bing. 2023.
\newblock \href {http://arxiv.org/abs/2305.15005} {Sentiment analysis in the era of large language models: A reality check}.

\bibitem[{Zhao et~al.(2024)Zhao, Wang, Abid, Angus, Garg, Kinnison, Sherstinsky, Molino, Addair, and Rishi}]{zhao2024loraland310finetuned}
Justin Zhao, Timothy Wang, Wael Abid, Geoffrey Angus, Arnav Garg, Jeffery Kinnison, Alex Sherstinsky, Piero Molino, Travis Addair, and Devvret Rishi. 2024.
\newblock \href {http://arxiv.org/abs/2405.00732} {Lora land: 310 fine-tuned llms that rival gpt-4, a technical report}.

\bibitem[{Zhao et~al.(2023)Zhao, Zhao, Lu, Wang, Tong, and Qin}]{chatgpt-emotional-dialogue}
Weixiang Zhao, Yanyan Zhao, Xin Lu, Shilong Wang, Yanpeng Tong, and Bing Qin. 2023.
\newblock \href {http://arxiv.org/abs/2304.09582} {Is chatgpt equipped with emotional dialogue capabilities?}

\end{thebibliography}

\appendix
\newpage
\onecolumn

\setcounter{table}{0}
\counterwithin{figure}{section}
\renewcommand{\thetable}{\Alph{section}\arabic{table}}

\section{Detailed Training Configurations}
\label{sec:detailed-training-config}
\subsection{Task-Specific Fine-tuning}
\label{sec:task-specific-fine-tuning-config}
For all model fine-tuning, the learning rate was 3e-5. The batch size was 2 with a gradient accumulation step of 4. We used a cosinusoidal learning rate scheduler without warming up. We applied a weight decay of 0.01 on all model parameters except for the biases and layer normalisation weights. 
For LLaMA-7B, Alpaca-7B, and LLaMA-3-8B, we stored model parameters in IEEE 754 half-precision float point format. For GPT-2, we stored the model parameters in standard single-precision floating-point format and did not apply LoRA during the fine-tuning. We followed the default LoRA configuration provided in Huggingface PEFT library \citep{peft}. We used the model perplexity on the development set as the early-stopping criterion. For EmoWOZ, we used the official development set. For IEMOCAP, when we performed the leave-one-session-out training, 10\% of the training data were randomly sampled as the development set. We applied stratified sampling based on the emotion labels. All open-source models were trained on a single Nvidia A100 40GB Graphics Card.

\subsection{ASR Simulation for EmoWOZ}
\label{sec:asr-simulator}
We fine-tuned a LLaMA-7B model using LoRA following configurations specified in Section \ref{sec:task-specific-fine-tuning} and \ref{sec:task-specific-fine-tuning-config} for one epoch on all IEMOCAP utterances.
The source was each of the IEMOCAP utterance transcription and the target was the corresponding OpenAI Whisper-medium hypothesis. We utilised a prompt template that formatted the source and target in natural language would best exploit the language modelling capability of the model:

\texttt{After adding automatic speech recognition errors, [SOURCE] becomes [TARGET]}


\section{Detailed Experimental Results}
\label{sec:detailed-results}

\begin{table*}[h]
\centering
\tiny
\begin{tabular}{llllllllll}
\toprule
Model & N & P & M & Neutral & Happy & Angry & Sad & WA & UA \\ \midrule
GPT-2 & 0 & 0\% & T & 0.7 (60.0/0.4) & 32.3 (43.6/25.6) & 35.3 (22.0/90.6) & 0.5 (30.0/0.3)  & 25.8 & 29.2\\
GPT-2 & 1 & 0\% & T & 10.9 (43.4/6.2) & 9.1 (62.0/4.9) & 29.0 (21.8/43.6) & 33.3 (22.8/61.9)  & 24.2 & 29.2 \\
GPT-2 & 0 & 25\% & T & 47.2 (30.9/100.0) & 0.0 (0.0/0.0) & 0.0 (0.0/0.0) & 0.0 (0.0/0.0)  & 30.9 & 25.0 \\ 
GPT-2 & 0 & 50\% & T & 47.2 (30.9/100.0) & 0.0 (0.0/0.0) & 0.0 (0.0/0.0) & 0.0 (0.0/0.0) & 30.9 & 25.0 \\ 
GPT-2 & 0 & 75\% & T & 47.2 (30.9/100.0) & 0.0 (0.0/0.0) & 0.0 (0.0/0.0) & 0.0 (0.0/0.0)  & 30.9 & 25.0 \\ 
GPT-2 & 0 & 100\% & T & 47.2 (30.9/100.0) & 0.0 (0.0/0.0) & 0.0 (0.0/0.0) & 0.0 (0.0/0.0)  & 30.9 & 25.0 \\ \midrule
LLaMA-7B & 0 & 0\% & T & 48.6 (37.5/69.3) & 21.8 (82.3/12.5) & 53.3 (40.8/76.8) & 6.9 (78.0/3.6)  & 41.1 & 40.5 \\
LLaMA-7B & 0 & 0\% & A & 50.3 (37.0/78.8) & 14.8 (79.3/8.2) & 54.7 (44.9/70.0) & 0.6 (100.0/0.3)  & 40.8 & 39.3 \\
LLaMA-7B & 1 & 0\% & T & 55.3 (42.9/77.5) & 56.2 (77.3/44.2) & 62.0 (55.0/71.0) & 11.1 (73.0/6.0)  & 52.3 & 49.7 \\
LLaMA-7B & 3 & 0\% & T & 54.2 (39.4/86.7) & 1.2 (90.9/0.6) & 44.1 (87.6/29.5) & 44.5 (39.5/50.8)  & 42.8 & 41.9 \\
LLaMA-7B & 0 & 25\% & T & 65.1 (64.6/65.6) & 77.0 (80.8/73.5) & 73.5 (72.6/74.4) & 74.2 (71.3/77.3)  & 72.0 & 72.7 \\
LLaMA-7B & 0 & 50\% & T & 69.1 (69.9/68.3) & 80.7 (80.6/80.7) & 77.3 (78.7/76.0) & 78.1 (75.6/80.8)  & 76.0 & 76.4 \\
LLaMA-7B & 0 & 75\% & T & 70.7 (67.3/74.5) & 82.2 (84.6/80.0) & 76.3 (80.5/72.4) & 78.1 (77.5/78.6)  & 76.5 & 76.4 \\
LLaMA-7B & 0 & 100\% & T & 69.7 (66.2/73.5) & 82.0 (82.4/81.7) & 79.0 (81.0/77.1) & 75.8 (80.6/71.6) & 76.3 & 76.0 \\ \midrule
Alpaca-7B & 0 & 0\% & T & 34.4 (49.6/26.3) & 62.8 (79.3/52.0) & 50.2 (34.2/94.6) & 44.5 (70.3/32.6)  & 48.8 & 51.4 \\
Alpaca-7B & 0 & 0\% & A & 37.0 (52.5/28.6) & 60.7 (75.3/50.8) & 48.5 (32.8/93.3) & 38.8 (77.6/25.8)  & 47.5 & 49.6 \\
Alpaca-7B & 1 & 0\% & T & 53.8 (49.4/59.0) & 59.5 (83.1/46.3) & 57.9 (43.5/86.9) & 37.0 (72.3/24.8)  & 54.1 & 54.3 \\
Alpaca-7B & 3 & 0\% & T & 55.4 (43.9/74.8) & 28.0 (90.6/16.6) & 64.5 (60.8/68.6) & 54.9 (55.1/54.7)  & 52.4 & 53.7 \\
Alpaca-7B & 0 & 25\% & T & 65.1 (67.3/63.0) & 77.0 (77.8/76.2) & 74.6 (72.1/77.2) & 70.8 (68.8/73.0)  & 71.7 & 72.4 \\ 
Alpaca-7B & 0 & 50\% & T & 69.1 (69.8/68.3) & 80.5 (78.7/82.4) & 78.3 (79.7/77.1) & 75.7 (75.9/75.6)  & 75.6 & 75.8 \\ 
Alpaca-7B & 0 & 75\% & T & 70.5 (66.3/75.4) & 80.6 (85.2/76.4) & 76.3 (78.7/74.2) & 74.2 (74.2/74.2)  & 75.2 & 75.0 \\ 
Alpaca-7B & 0 & 100\% & T & 69.3 (69.1/69.5) & 81.0 (82.3/79.8) & 78.8 (79.2/78.4) & 76.5 (74.7/78.3)  & 76.0 & 76.5 \\ \midrule
LLaMA-3-8B & 0 & 0\% & T & 3.4 (55.6/1.8) & 55.9 (42.7/81.0) & 51.0 (38.7/75.0) & 19.9 (55.7/12.1) & 41.8 & 42.5 \\
LLaMA-3-8B & 0 & 0\% & A & 2.1 (56.2/1.1) & 51.2 (36.1/88.3) & 52.3 (47.4/58.5) & 14.3 (65.4/8.0) & 39.7 & 39.0 \\
LLaMA-3-8B & 1 & 0\% & T & 52.7 (50.1/55.6) & 67.1 (75.9/60.1) & 60.4 (46.7/85.4) & 37.9 (82.9/24.5) & 56.8 & 56.4 \\
LLaMA-3-8B & 3 & 0\% & T & 35.8 (62.0/25.2) & 67.0 (79.5/57.9) & 63.4 (53.3/78.2) & 60.0 (46.1/86.2) & 57.4 & 61.9 \\
LLaMA-3-8B & 0 & 25\% & T & 68.2 (66.8/69.6) & 78.5 (76.6/80.5) & 74.4 (76.0/72.9) & 72.4 (76.4/68.7) & 73.3 & 72.9 \\ 
LLaMA-3-8B & 0 & 50\% & T & 69.7 (71.0/68.5) & 81.4 (79.0/84.0) & 77.6 (82.3/73.3) & 77.4 (74.8/80.2) & 76.3 & 76.5 \\ 
LLaMA-3-8B & 0 & 75\% & T & 71.8 (73.8/70.0) & 83.2 (81.7/84.7) & 80.6 (82.2/79.1) & 80.2 (77.8/82.7) & 78.7 & 79.1 \\ 
LLaMA-3-8B & 0 & 100\% & T & 73.2 (74.0/72.5) & 84.3 (83.0/85.8) & 81.6 (83.1/80.2) & 81.0 (80.4/81.6) & 79.8 & 80.0 \\ \midrule
GPT-3.5 & 0 & 0\% & T & 51.6 (35.1/97.3) & 28.5 (90.5/16.9) & 31.7 (79.9/19.8) & 27.1 (81.5/16.2) & 42.2 & 37.6 \\
GPT-3.5 & 0 & 0\% & A & 51.3 (34.9/96.5) & 31.7 (88.8/19.3) & 33.0 (83.2/20.6) & 23.2 (80.8/13.6)  & 42.3 & 37.5 \\
GPT-3.5 & 1 & 0\% & T & 57.7 (42.7/88.9) & 59.4 (84.8/45.7) & 56.6 (76.1/45.1) & 45.9 (80.0/32.2)  & 56.3 & 53.0 \\
GPT-3.5 & 3 & 0\% & T & 60.1 (48.9/78.2) & 66.0 (80.0/56.1) & 63.8 (76.4/54.8) & 59.4 (67.1/53.3)  & 62.1 & 60.6 \\ \midrule
GPT-4 & 0 & 0\% & T & 51.7 (35.0/99.3) & 28.9 (97.5/16.9) & 29.4 (95.5/17.4) & 28.2 (91.4/16.7) & 42.4 & 37.6 \\ 
GPT-4 & 0 & 0\% & A & 51.5 (34.8/98.9) & 27.6 (94.3/16.2) & 30.4 (95.7/18.0) & 25.9 (89.6/15.1)  & 41.9 & 37.1 \\
GPT-4 & 1 & 0\% & T & 62.3 (48.1/88.2) & 59.7 (83.4/46.5) & 70.4 (81.6/61.9) & 60.9 (81.1/48.7)  & 62.9 & 61.3 \\
GPT-4 & 3 & 0\% & T & 61.6 (49.0/83.1) & 60.7 (84.9/47.2) & 67.6 (85.2/56.1) & 68.4 (71.6/65.4)  & 63.8 & 63.0 \\
\bottomrule
\end{tabular}
\caption{F1(precision/recall), UA, and WA of LLMs on IEMOCAP under the 4-Way classification set-up. In table headers, ``N'' stands for the number of ICL samples in the prompt; ``P'' stands for the proportion of training data used for fine-tuning; ``M'' stands for the modality of input, either transcription (T) or ASR hypothesis (A). \label{tab:iemocap-4way-full}}
\end{table*}


\begin{table*}[h]
\centering
\tiny
\setlength\tabcolsep{3pt}
\begin{tabular}{lllllllllll}
\toprule
Model & N & P & M & Neutral & Happy & Angry & Sad & Other & WA & UA \\ \midrule
GPT-2 & 0 & 0\% & T & 0.1 (7.7/0.1) & 23.1 (35.2/17.2) & 27.5 (16.5/81.1) & 0.5 (42.9/0.3) & 15.2 (19.3/12.5)  & 19.0 & 22.3 \\
GPT-2 & 1 & 0\% & T & 13.3 (31.6/8.4) & 26.6 (35.4/21.3) & 19.2 (16.3/23.5) & 27.4 (17.0/70.5) & 0.0 (0.0/0.0)  & 20.1 & 24.7 \\
GPT-2 & 0 & 25\% & T & 37.0 (22.7/100.0) & 0.0 (0.0/0.0) & 0.0 (0.0/0.0) & 0.0 (0.0/0.0) & 0.0 (0.0/0.0)  & 22.7 & 20.0 \\
GPT-2 & 0 & 50\% & T & 37.0 (22.7/100.0) & 0.0 (0.0/0.0) & 0.0 (0.0/0.0) & 0.0 (0.0/0.0) & 0.0 (0.0/0.0)  & 22.7 & 20.0 \\
GPT-2 & 0 & 75\% & T & 37.0 (22.7/100.0) & 0.0 (0.0/0.0) & 0.0 (0.0/0.0) & 0.0 (0.0/0.0) & 0.0 (0.0/0.0)  & 22.7 & 20.0 \\
GPT-2 & 0 & 100\% & T & 37.0 (22.7/100.0) & 0.0 (0.0/0.0) & 0.0 (0.0/0.0) & 0.0 (0.0/0.0) & 0.0 (0.0/0.0)  & 22.7 & 20.0 \\ \midrule

LLaMA-7B & 0 & 0\% & T & 38.7 (29.8/55.2) & 21.5 (81.8/12.3) & 37.8 (30.3/50.0) & 6.3 (65.5/3.3) & 44.5 (42.1/47.3)  & 35.6 & 33.6 \\
LLaMA-7B & 0 & 0\% & A & 37.0 (30.9/46.3) & 13.4 (80.5/7.3) & 17.3 (40.5/11.0) & 0.0 (0.0/0.0) & 48.0 (34.6/78.2)  & 34.5 & 28.6 \\
LLaMA-7B & 1 & 0\% & T & 2.2 (59.4/1.1) & 0.1 (100.0/0.1) & 15.1 (60.9/8.6) & 0.0 (0.0/0.0) & 41.6 (26.5/97.2)  & 27.3 & 21.4 \\
LLaMA-7B & 3 & 0\% & T & 11.1 (22.4/7.4) & 0.2 (16.7/0.1) & 0.0 (0.0/0.0) & 0.0 (0.0/0.0) & 41.2 (26.5/92.3)  & 26.2 & 20.0 \\
LLaMA-7B & 0 & 25\% & T & 51.2 (48.1/54.9) & 74.3 (72.4/76.3) & 49.3 (54.6/44.9) & 62.5 (63.3/61.6) & 49.9 (51.3/48.6)  & 57.4 & 57.3 \\
LLaMA-7B & 0 & 50\% & T & 57.9 (54.5/61.7) & 79.0 (77.1/80.9) & 54.7 (57.7/52.0) & 69.3 (73.8/65.3) & 55.6 (56.7/54.6)  & 63.1 & 62.9 \\
LLaMA-7B & 0 & 75\% & T & 60.8 (58.2/63.7) & 82.3 (82.2/82.5) & 56.4 (60.9/52.5) & 72.9 (68.9/77.4) & 57.3 (59.6/55.2)  & 65.9 & 66.3 \\
LLaMA-7B & 0 & 100\% & T & 53.9 (57.5/50.6) & 80.9 (77.6/84.5) & 55.8 (63.6/49.8) & 72.1 (69.8/74.5) & 57.4 (54.5/60.7)  & 64.0 & 64.0 \\ \midrule

Alpaca-7B & 0 & 0\% & T & 18.1 (42.5/11.5) & 52.6 (78.3/39.6) & 29.2 (34.9/25.0) & 29.2 (69.4/18.5) & 48.4 (33.6/86.6)  & 40.5 & 36.2 \\
Alpaca-7B & 0 & 0\% & A & 15.8 (41.3/9.8) & 49.5 (74.3/37.1) & 23.4 (31.5/18.6) & 21.8 (72.8/12.8) & 48.2 (32.9/90.0)  & 38.7 & 33.6 \\
Alpaca-7B & 1 & 0\% & T & 0.2 (28.6/0.1) & 0.5 (100.0/0.2) & 5.1 (74.4/2.6) & 0.0 (0.0/0.0) & 42.0 (26.6/99.5)  & 26.9 & 20.5 \\
Alpaca-7B & 3 & 0\% & T & 9.4 (17.1/6.4) & 3.7 (24.3/2.0) & 3.1 (18.1/1.7) & 5.7 (10.4/3.9) & 39.6 (26.1/81.5)  & 24.4 & 19.1 \\
Alpaca-7B & 0 & 25\% & T & 48.9 (53.2/45.3) & 73.8 (68.0/80.8) & 51.2 (56.4/46.9) & 63.4 (60.5/66.5) & 52.1 (51.8/52.4)  & 58.2 & 58.4 \\ 
Alpaca-7B & 0 & 50\% & T & 56.4 (54.0/59.1) & 78.7 (76.8/80.6) & 55.5 (58.9/52.4) & 68.8 (74.6/63.8) & 57.4 (57.0/57.9)  & 63.2 & 62.8 \\ 
Alpaca-7B & 0 & 75\% & T & 57.9 (59.5/56.4) & 81.0 (77.5/84.9) & 58.9 (62.5/55.7) & 70.2 (64.8/76.7) & 56.9 (59.2/54.9)  & 65.0 & 65.7 \\ 
Alpaca-7B & 0 & 100\% & T & 54.7 (55.6/53.8) & 81.1 (77.7/84.8) & 58.0 (60.3/55.9) & 69.8 (65.3/75.0) & 55.1 (57.8/52.6)  & 63.6 & 64.4 \\ \midrule

LLaMA-3-8B & 0 & 0\% & T & 1.6 (46.7/0.8) & 44.5 (30.1/85.3) & 38.5 (27.3/65.1) & 12.4 (40.9/7.3) & 0.4 (10.3/0.2) & 29.4 & 31.7 \\
LLaMA-3-8B & 0 & 0\% & A & 0.9 (44.4/0.5) & 41.0 (26.5/90.7) & 39.5 (32.4/50.6) & 9.2 (47.8/5.1) & 1.6 (23.6/0.8) & 28.2 & 29.5 \\ 
LLaMA-3-8B & 1 & 0\% & T & 44.1 (43.5/44.8) & 68.5 (66.7/70.4) & 39.8 (25.8/87.5) & 22.8 (85.1/13.2) & 2.7 (21.6/1.4) & 40.5 & 43.5 \\
LLaMA-3-8B & 3 & 0\% & T & 23.1 (63.9/14.1) & 64.2 (70.8/58.7) & 44.7 (40.1/50.5) & 37.6 (23.5/94.7) & 0.7 (21.2/0.3) & 37.1 & 43.7 \\
LLaMA-3-8B & 0 & 25\% & T & 54.6 (54.9/54.4) & 76.6 (74.4/78.9) & 46.9 (62.4/37.6) & 67.4 (65.9/69.1) & 54.4 (50.8/58.5) & 60.5 & 59.7 \\ 
LLaMA-3-8B & 0 & 50\% & T & 58.4 (60.3/56.7) & 80.4 (78.3/82.7) & 54.4 (65.5/46.5) & 72.8 (70.5/75.3) & 60.7 (57.2/64.6) & 65.6 & 65.1 \\ 
LLaMA-3-8B & 0 & 75\% & T & 60.2 (61.5/59.0) & 81.9 (78.6/85.4) & 56.9 (68.2/48.9) & 74.6 (73.8/75.4) & 61.8 (58.7/65.1) & 67.2 & 66.8 \\ 
LLaMA-3-8B & 0 & 100\% & T & 63.9 (66.1/61.9) & 83.2 (81.4/85.1) & 59.5 (66.0/54.1) & 75.8 (76.7/74.9) & 63.4 (59.8/67.5) & 69.5 & 68.7 \\ \midrule

GPT-3.5 & 0 & 0\% & T & 43.6 (28.7/91.2) & 29.2 (87.0/17.5) & 29.4 (63.4/19.1) & 26.3 (72.2/16.1) & 39.2 (52.5/31.3)  & 37.9 & 35.1 \\
GPT-3.5 & 0 & 0\% & A & 43.7 (28.8/90.9) & 32.8 (87.3/20.2) & 29.2 (61.1/19.2) & 24.6 (75.7/14.7) & 38.4 (51.1/30.7)  & 38.1 & 35.1 \\
GPT-3.5 & 1 & 0\% & T & 45.9 (36.3/62.5) & 63.3 (74.8/54.9) & 49.8 (46.7/53.4) & 48.8 (65.4/38.9) & 38.0 (44.7/33.1)  & 48.3 & 48.6 \\
GPT-3.5 & 3 & 0\% & T & 47.4 (43.1/52.7) & 67.6 (69.8/65.6) & 49.5 (40.4/63.9) & 54.1 (45.4/66.8) & 18.3 (41.4/11.7)  & 48.3 & 52.1 \\ \midrule

GPT-4 & 0 & 0\% & T & 43.1 (28.0/93.7) & 28.1 (94.1/16.5) & 27.4 (82.6/16.4) & 29.4 (85.7/17.7) & 37.8 (53.8/29.1)  & 37.5 & 34.7 \\
GPT-4 & 0 & 0\% & A & 42.9 (27.7/95.3) & 27.6 (94.3/16.1) & 30.7 (79.2/19.0) & 27.2 (79.5/16.4) & 31.8 (51.5/23.0)  & 36.4 & 34.0 \\
GPT-4 & 1 & 0\% & T & 51.1 (37.9/78.2) & 58.9 (80.4/46.5) & 55.3 (49.9/61.9) & 54.5 (62.0/48.7) & 27.0 (45.6/19.1)  & 49.0 & 50.9 \\
GPT-4 & 3 & 0\% & T & 49.6 (39.8/65.8) & 59.9 (81.7/47.2) & 54.7 (53.5/56.1) & 58.3 (52.5/65.4) & 30.9 (40.5/24.9)  & 49.4 & 51.9 \\
\bottomrule
\end{tabular}
\caption{F1(precision/recall), UA, and WA of LLMs on IEMOCAP under the 5-Way classification set-up. In table headers, ``N'' stands for the number of ICL samples in the prompt; ``P'' stands for the proportion of training data used for fine-tuning; ``M'' stands for the modality of input, either transcription (T) or ASR hypothesis (A). \label{tab:iemocap-5way-full}}
\end{table*}

\begin{table*}[h]
\centering
\setlength\tabcolsep{2.5pt}
\tiny
\begin{tabular}{lllllllllllll}
\toprule
Model & N & P & M & Neutral & Fearful & Dissatisfied & Apologetic & Abusive & Excited & Satisfied & MF1 & WF1 \\ \midrule
GPT-2 & 0 & 0\% & T & 0.1 (100.0/0.0) & 0.0 (0.0/0.0) & 9.3 (5.6/27.8) & 0.0 (0.0/0.0) & 0.0 (0.0/0.0) & 2.8 (1.4/64.8) & 31.4 (35.7/28.1)  & 7.3 & 24.0 \\
GPT-2 & 1 & 0\% & T & 81.2 (69.8/97.0) & 0.0 (0.0/0.0) & 0.0 (0.0/0.0) & 23.3 (14.8/54.8) & 0.0 (0.0/0.0) & 0.0 (0.0/0.0) & 0.0 (0.0/0.0)  & 3.9 & 0.6 \\
GPT-2 & 0 & 25\% & T & 82.4 (70.1/99.8) & 0.0 (0.0/0.0) & 0.0 (0.0/0.0) & 69.9 (71.4/68.5) & 0.0 (0.0/0.0) & 0.0 (0.0/0.0) & 0.0 (0.0/0.0)  & 11.7 & 1.9 \\
GPT-2 & 0 & 50\% & T & 82.3 (70.0/100.0) & 0.0 (0.0/0.0) & 0.0 (0.0/0.0) & 69.6 (95.2/54.8) & 0.0 (0.0/0.0) & 0.0 (0.0/0.0) & 0.0 (0.0/0.0)  & 11.6 & 1.9 \\
GPT-2 & 0 & 75\% & T & 82.4 (70.3/99.5) & 0.0 (0.0/0.0) & 0.0 (0.0/0.0) & 58.9 (47.9/76.7) & 0.0 (0.0/0.0) & 0.0 (0.0/0.0) & 0.0 (0.0/0.0)  & 9.8 & 1.6 \\
GPT-2 & 0 & 100\% & T & 82.3 (70.0/99.8) & 0.0 (0.0/0.0) & 0.0 (0.0/0.0) & 68.1 (74.2/63.0) & 0.0 (0.0/0.0) & 0.0 (0.0/0.0) & 0.0 (0.0/0.0)  & 11.4 & 1.9 \\ \midrule
LLaMA-7B & 0 & 0\% & T & 82.1 (69.7/100.0) & 0.0 (0.0/0.0) & 0.3 (33.3/0.2) & 0.0 (0.0/0.0) & 0.0 (0.0/0.0) & 6.3 (75.0/3.3) & 0.0 (0.0/0.0)  & 1.1 & 0.3 \\
LLaMA-7B & 0 & 0\% & A & 82.1 (69.7/100.0) & 0.0 (0.0/0.0) & 0.0 (0.0/0.0) & 0.0 (0.0/0.0) & 0.0 (0.0/0.0) & 0.0 (0.0/0.0) & 0.0 (0.0/0.0)  & 0.0 & 0.0 \\
LLaMA-7B & 1 & 0\% & T & 83.0 (78.1/88.5) & 26.1 (60.0/16.7) & 2.6 (47.1/1.3) & 0.0 (0.0/0.0) & 57.9 (52.4/64.7) & 16.0 (9.2/58.2) & 59.0 (74.1/49.0)  & 26.9 & 42.6 \\
LLaMA-7B & 3 & 0\% & T & 27.9 (81.2/16.9) & 0.0 (0.0/0.0) & 0.0 (0.0/0.0) & 0.0 (0.0/0.0) & 0.0 (0.0/0.0) & 0.0 (0.0/0.0) & 39.2 (24.4/99.2)  & 6.5 & 27.2 \\
LLaMA-7B & 0 & 25\% & T & 93.9 (91.5/96.4) & 26.1 (60.0/16.7) & 55.2 (81.6/41.7) & 72.3 (93.5/58.9) & 11.1 (100.0/5.9) & 43.6 (69.0/31.9) & 90.9 (89.1/92.7)  & 49.9 & 79.5 \\
LLaMA-7B & 0 & 50\% & T & 94.4 (93.1/95.8) & 41.7 (83.3/27.8) & 68.6 (80.0/60.1) & 75.8 (92.2/64.4) & 64.0 (100.0/47.1) & 51.4 (69.8/40.7) & 91.1 (89.8/92.5)  & 65.4 & 83.6 \\
LLaMA-7B & 0 & 75\% & T & 93.8 (93.0/94.5) & 35.3 (37.5/33.3) & 61.7 (84.9/48.5) & 57.4 (41.5/93.2) & 69.2 (100.0/52.9) & 50.6 (54.4/47.3) & 90.8 (88.7/93.1)  & 60.8 & 81.3 \\
LLaMA-7B & 0 & 100\% & T & 94.2 (93.3/95.2) & 43.8 (50.0/38.9) & 68.1 (78.3/60.3) & 75.4 (93.9/63.0) & 69.2 (100.0/52.9) & 51.3 (63.9/42.9) & 90.7 (88.8/92.6)  & 66.4 & 83.2 \\ \midrule

Alpaca-7B & 0 & 0\% & T & 65.4 (85.3/53.1) & 1.9 (1.1/11.1) & 24.8 (28.5/22.0) & 46.0 (85.2/31.5) & 0.0 (0.0/0.0) & 18.1 (13.8/26.4) & 53.3 (38.7/85.7)  & 24.0 & 44.6 \\
Alpaca-7B & 0 & 0\% & A & 65.5 (83.0/54.1) & 2.0 (1.1/11.1) & 22.9 (26.2/20.4) & 52.9 (93.1/37.0) & 0.0 (0.0/0.0) & 17.3 (15.4/19.8) & 50.7 (37.2/80.0)  & 24.3 & 42.6 \\ 
Alpaca-7B & 1 & 0\% & T & 75.4 (81.4/70.2) & 3.4 (1.9/22.2) & 13.8 (23.8/9.8) & 26.2 (100.0/15.1) & 30.0 (100.0/17.6) & 8.9 (4.8/65.9) & 67.4 (69.4/65.6)  & 25.0 & 51.2 \\
Alpaca-7B & 3 & 0\% & T & 65.4 (85.3/53.1) & 1.9 (1.1/11.1) & 24.8 (28.5/22.0) & 46.0 (85.2/31.5) & 0.0 (0.0/0.0) & 18.1 (13.8/26.4) & 53.3 (38.7/85.7)  & 24.0 & 44.6 \\
Alpaca-7B & 0 & 25\% & T & 93.3 (91.9/94.7) & 17.4 (40.0/11.1) & 53.5 (72.5/42.4) & 74.8 (92.0/63.0) & 0.0 (0.0/0.0) & 45.4 (64.0/35.2) & 90.2 (86.6/94.2)  & 46.9 & 78.7 \\ 
Alpaca-7B & 0 & 50\% & T & 94.4 (93.1/95.8) & 43.5 (100.0/27.8) & 68.1 (79.9/59.3) & 74.2 (83.1/67.1) & 64.0 (100.0/47.1) & 46.0 (66.7/35.2) & 91.0 (89.6/92.6)  & 64.5 & 83.2 \\ 
Alpaca-7B & 0 & 75\% & T & 93.6 (91.0/96.4) & 35.7 (50.0/27.8) & 45.6 (90.6/30.5) & 75.7 (79.1/72.6) & 38.1 (100.0/23.5) & 50.7 (67.3/40.7) & 91.1 (88.7/93.6)  & 56.1 & 78.0 \\ 
Alpaca-7B & 0 & 100\% & T & 94.0 (92.0/96.1) & 10.5 (100.0/5.6) & 62.2 (76.4/52.5) & 71.8 (73.9/69.9) & 0.0 (0.0/0.0) & 39.3 (77.4/26.4) & 90.7 (90.2/91.1)  & 45.8 & 80.7 \\ \midrule

LLaMA-3-8B & 0 & 0\% & T & 44.2 (79.5/30.6) & 1.3 (0.6/55.6) & 1.0 (13.0/0.5) & 24.0 (14.6/67.1) & 26.8 (15.8/88.2) & 5.7 (2.9/81.3) & 59.3 (59.6/59.1) & 19.7 & 42.4 \\
LLaMA-3-8B & 0 & 0\% & A & 47.0 (80.7/33.1) & 1.2 (0.6/50.0) & 1.0 (12.5/0.5) & 24.1 (14.7/67.1) & 27.5 (16.3/88.2) & 5.7 (3.0/78.0) & 59.2 (59.3/59.1) & 19.8 & 42.3 \\
LLaMA-3-8B & 1 & 0\% & T & 83.5 (76.3/92.3) & 4.5 (3.8/5.6) & 2.4 (16.0/1.3) & 39.4 (30.8/54.8) & 9.0 (4.7/100.0) & 35.4 (46.4/28.6) & 50.5 (87.4/35.5) & 23.5 & 38.0 \\
LLaMA-3-8B & 3 & 0\% & T & 55.5 (85.1/41.1) & 0.0 (0.0/0.0) & 0.3 (20.0/0.2) & 35.3 (62.1/24.7) & 2.3 (1.2/100.0) & 24.6 (60.9/15.4) & 54.8 (39.2/90.9) & 19.5 & 39.9 \\
LLaMA-3-8B & 0 & 25\% & T & 93.6 (90.7/96.7) & 27.3 (75.0/16.7) & 52.4 (87.6/37.4) & 74.6 (97.8/60.3) & 74.1 (100.0/58.8) & 49.0 (64.3/39.6) & 90.4 (89.5/91.3) & 61.3 & 79.2 \\
LLaMA-3-8B & 0 & 50\% & T & 93.8 (91.1/96.7) & 26.1 (60.0/16.7) & 56.5 (86.1/42.1) & 76.4 (94.0/64.4) & 74.1 (100.0/58.8) & 49.3 (62.7/40.7) & 90.5 (90.1/91.0) & 62.2 & 80.3 \\ 
LLaMA-3-8B & 0 & 75\% & T & 94.3 (92.4/96.3) & 38.5 (62.5/27.8) & 64.0 (85.2/51.3) & 75.2 (83.3/68.5) & 64.0 (100.0/47.1) & 48.6 (66.0/38.5) & 90.8 (89.5/92.2) & 63.5 & 82.2 \\ 
LLaMA-3-8B & 0 & 100\% & T & 94.5 (92.4/96.7) & 50.0 (100.0/33.3) & 66.2 (85.2/54.1) & 74.4 (85.7/65.8) & 78.6 (100.0/64.7) & 52.1 (69.1/41.8) & 90.8 (90.5/91.0) & 68.7 & 82.9 \\ \midrule

GPT-3.5 & 0 & 0\% & T & 82.8 (76.9/89.8) & 20.7 (27.3/16.7) & 8.2 (35.0/4.6) & 61.9 (87.5/47.9) & 61.5 (88.9/47.1) & 31.6 (27.4/37.4) & 50.0 (58.9/43.5)  & 39.0 & 40.0 \\
GPT-3.5 & 0 & 0\% & A & 82.9 (76.8/90.0) & 28.6 (100.0/16.7) & 8.2 (35.4/4.6) & 61.9 (87.5/47.9) & 61.5 (88.9/47.1) & 31.5 (27.2/37.4) & 49.7 (58.8/43.1)  & 40.2 & 39.8 \\
GPT-3.5 & 1 & 0\% & T & 66.0 (82.3/55.1) & 36.4 (100.0/22.2) & 13.5 (33.1/8.4) & 7.9 (100.0/4.1) & 75.7 (70.0/82.4) & 10.7 (5.9/61.5) & 56.0 (42.6/81.6)  & 33.3 & 43.2 \\
GPT-3.5 & 3 & 0\% & T & 57.9 (82.1/44.7) & 10.0 (50.0/5.6) & 16.6 (34.4/10.9) & 36.0 (100.0/21.9) & 69.0 (83.3/58.8) & 6.4 (3.4/71.4) & 59.3 (46.5/81.8)  & 32.9 & 46.7 \\ \midrule

GPT-4 & 0 & 0\% & T & 88.3 (86.0/90.8) & 50.0 (100.0/33.3) & 16.4 (47.2/9.9) & 52.5 (37.4/87.7) & 74.1 (100.0/58.8) & 42.2 (36.2/50.5) & 79.0 (78.5/79.6)  & 52.4 & 62.3 \\
GPT-4 & 0 & 0\% & A & 88.3 (82.7/94.6) & 41.7 (83.3/27.8) & 47.9 (70.1/36.4) & 47.9 (33.2/86.3) & 75.9 (91.7/64.7) & 39.8 (31.6/53.8) & 66.6 (89.4/53.1) & 53.3 & 60.8 \\
GPT-4 & 1 & 0\% & T & 78.8 (93.8/68.0) & 41.7 (83.3/27.8) & 52.5 (46.7/60.1) & 42.7 (28.2/87.7) & 83.3 (78.9/88.2) & 14.6 (8.0/80.2) & 71.8 (63.8/82.1)  & 51.1 & 64.4 \\
GPT-4 & 3 & 0\% & T & 83.2 (91.9/76.0) & 26.1 (60.0/16.7) & 51.1 (48.0/54.6) & 55.0 (42.0/79.5) & 77.8 (73.7/82.4) & 28.0 (20.3/45.1) & 74.2 (63.0/90.4)  & 52.0 & 66.5 \\
\bottomrule
\end{tabular}
\caption{F1(precision/recall), MF1 and WF1 of LLMs on EmoWOZ. In table headers, ``N'' stands for the number of ICL samples in the prompt; ``P'' stands for the proportion of training data used for fine-tuning; ``M'' stands for the modality of input, either transcription (T) or ASR hypothesis (A). \label{tab:emowoz-full}}
\end{table*}


\begin{table*}[h]
\centering
\tiny
\begin{tabular}{llllcccccc}
\toprule
\multirow{2}{*}{Model} & \multirow{2}{*}{N} & \multirow{2}{*}{P} & \multirow{2}{*}{M} & \multicolumn{2}{c}{Development Set} & \multicolumn{2}{c}{Test Set} \\
  &  &  &  & Depressed & Not Depressed & Depressed & Not Depressed \\ \midrule
GPT-2 & 0 & 0\% & T &  0.0 (0.0/0.0) & 82.5 (70.2/100.0) & 0.0 (0.0/0.0) & 79.3 (65.7/100.0) \\ \midrule

LLaMA-7B & 0 & 0\% & T & 47.5 (31.1/100.0) & 11.4 (100.0/6.1) & 52.2 (35.3/100.0) & 8.3 (100.0/4.3) \\
LLaMA-7B & 0 & 0\% & A & 45.9 (29.8/100.0) & 0.0 (0.0/0.0) & 51.1 (34.3/100.0) & 0.0 (0.0/0.0) & \\
LLaMA-7B & 1 & 0\% & T & 0.0 (0.0/0.0) & 82.5 (70.2/100.0) & 0.0 (0.0/0.0) & 79.3 (65.7/100.0) \\
LLaMA-7B & 3 & 0\% & T & 42.1 (27.9/85.7) & 10.8 (50.0/6.1) & 48.9 (33.3/91.7) & 8.0 (50.0/4.3) \\

LLaMA-7B & 0 & 25\% & T & 0.0 (0.0/0.0) & 81.0 (69.6/97.0) & 0.0 (0.0/0.0) & 79.3 (65.7/100.0) \\
LLaMA-7B & 0 & 50\% & T & 0.0 (0.0/0.0) & 81.0 (69.6/97.0) & 0.0 (0.0/0.0) & 79.3 (65.7/100.0 \\ 
LLaMA-7B & 0 & 75\% & T & 0.0 (0.0/0.0) & 79.5 (68.9/93.9) & 0.0 (0.0/0.0) & 79.3 (65.7/100.0) \\ 
LLaMA-7B & 0 & 100\% & T & 0.0 (0.0/0.0) & 76.3 (67.4/87.9) & 0.0 (0.0/0.0) & 79.3 (65.7/100.0) \\  \midrule

Alpaca-7B & 0 & 0\% & T & 47.5 (31.1/100.0) & 11.4 (100.0/6.1) & 53.3 (36.4/100.0) & 16.0 (100.0/8.7) \\
Alpaca-7B & 0 & 0\% & A & 45.9 (29.8/100.0) & 0.0 (0.0/0.0) & 53.3 (36.4/100.0) & 16.0 (100.0/8.7) & \\ 
Alpaca-7B & 1 & 0\% & T & 0.0 (0.0/0.0) & 82.5 (70.2/100.0) & 15.4 (100.0/8.3) & 80.7 (67.6/100.0) \\
Alpaca-7B & 3 & 0\% & T & 45.9 (29.8/100.0) & 0.0 (0.0/0.0) & 51.1 (34.3/100.0) & 0.0 (0.0/0.0) \\ 
Alpaca-7B & 0 & 25\% & T & 12.5 (50.0/7.1) & 82.1 (71.1/97.0) & 0.0 (0.0/0.0) & 77.2 (64.7/95.7) \\ 
Alpaca-7B & 0 & 50\% & T & 11.8 (33.3/7.1) & 80.5 (70.5/93.9) & 0.0 (0.0/0.0) & 77.2 (64.7/95.7) \\ 
Alpaca-7B & 0 & 75\% & T & 10.5 (20.0/7.1) & 77.3 (69.0/87.9) & 13.3 (33.3/8.3) & 76.4 (65.6/91.3) \\ 
Alpaca-7B & 0 & 100\% & T & 18.2 (25.0/14.3) & 75.0 (69.2/81.8) & 0.0 (0.0/0.0) & 75.0 (63.6/91.3)\\  \midrule

LLaMA-3-8B & 0 & 0\% & T & 47.1 (32.4/85.7) & 37.2 (80.0/24.2) & 43.2 (32.0/66.7) & 36.4 (60.0/26.1) \\
LLaMA-3-8B & 0 & 0\% & A & 46.4 (31.0/92.9) & 21.1 (80.0/12.1) & 42.9 (30.0/75.0) & 14.3 (40.0/8.7) \\
LLaMA-3-8B & 1 & 0\% & T & 0.0 (0.0/0.0) & 82.5 (70.2/100.0) & 0.0 (0.0/0.0) & 79.3 (65.7/100.0) \\
LLaMA-3-8B & 3 & 0\% & T & 0.0 (0.0/0.0) & 82.5 (70.2/100.0) & 0.0 (0.0/0.0) & 79.3 (65.7/100.0) \\ 
LLaMA-3-8B & 0 & 25\% & T & 27.0 (21.7/35.7) & 52.6 (62.5/45.5) & 29.6 (26.7/33.3) & 55.8 (60.0/52.2) \\ 
LLaMA-3-8B & 0 & 50\% & T & 0.0 (0.0/0.0) & 82.5 (70.2/100.0) & 0.0 (0.0/0.0) & 77.2 (64.7/95.7) \\ 
LLaMA-3-8B & 0 & 75\% & T & 0.0 (0.0/0.0) & 82.5 (70.2/100.0) & 47.1 (36.4/66.7) & 50.0 (69.2/39.1) \\ 
LLaMA-3-8B & 0 & 100\% & T & 20.0 (33.3/14.3) & 78.4 (70.7/87.9) & 47.1 (36.4/66.7) & 50.0 (69.2/39.1) \\  \midrule

GPT-3.5 & 0 & 0\% & T & 54.5 (60.0/50.0) & 79.2 (76.0/82.6) & 64.3 (64.3/64.3) & 84.8 (84.8/84.8)\\
GPT-3.5 & 0 & 0\% & A & 37.5 (75.0/25.0) & 81.5 (71.0/95.7) & 56.0 (63.6/50.0) & 84.1 (80.6/87.9)\\
GPT-3.5 & 1 & 0\% & T & 13.3 (33.3/8.3) & 76.4 (65.6/91.3) & 40.0 (45.5/35.7) & 78.3 (75.0/81.8)\\
GPT-3.5 & 3 & 0\% & T & 37.5 (75.0/25.0) & 81.5 (71.0/95.7) & 56.0 (63.6/50.0) & 84.1 (80.6/87.9)\\ \midrule

GPT-4 & 0 & 0\% & T & 63.6 (70.0/58.3) & 83.3 (80.0/87.0) & 59.3 (61.5/57.1) & 83.6 (82.4/84.8) \\
GPT-4 & 0 & 0\% & A & 44.4 (66.7/33.3) & 80.8 (72.4/91.3) & 41.7 (50.0/35.7) & 80.0 (75.7/84.8) \\
GPT-4 & 1 & 0\% & T & 80.0 (76.9/83.3) & 88.9 (90.9/87.0) & 55.6 (45.5/71.4) & 72.4 (84.0/63.6)\\
GPT-4 & 3 & 0\% & T & 74.1 (66.7/83.3) & 83.7 (90.0/78.3) & 58.5 (44.4/85.7) & 58.5 (44.4/85.7) \\

\bottomrule
\end{tabular}
\caption{F1(precision/recall) of LLMs on DAIC-WOZ. In table headers, ``N'' stands for the number of ICL samples in the prompt; ``P'' stands for the proportion of training data used for fine-tuning; ``M'' stands for the modality of input, either transcription (T) or ASR hypothesis (A). \label{tab:daicwoz-full}}
\end{table*}

\end{document}